\documentclass{article} 
\usepackage{iclr2025_conference,times}

\usepackage{amsmath,amsfonts,bm}

\def\eqref#1{equation~\ref{#1}}

\def\1{\bm{1}}

\DeclareMathAlphabet{\mathsfit}{\encodingdefault}{\sfdefault}{m}{sl}
\SetMathAlphabet{\mathsfit}{bold}{\encodingdefault}{\sfdefault}{bx}{n}

\usepackage{url}
\usepackage[utf8]{inputenc} 
\usepackage[T1]{fontenc}    
\usepackage[colorlinks]{hyperref}       
\usepackage{url}            
\usepackage{booktabs}       
\usepackage{amsfonts}       
\usepackage{nicefrac}       
\usepackage{microtype}      
\usepackage{xcolor}         
\usepackage{graphicx}
\usepackage{multirow}
\usepackage[font={small}]{caption}
\usepackage{subcaption}
\usepackage{xspace}
\usepackage{amsmath}
\usepackage{amssymb}
\usepackage{enumitem}
\usepackage{bbm}
\usepackage{pifont}
\usepackage[capitalize]{cleveref}
\usepackage{tabularx}

\let\oldding\ding
\renewcommand{\ding}[2][1]{\scalebox{#1}{\oldding{#2}}}

\usepackage[most]{tcolorbox}
\usepackage{color, colortbl}
\usepackage{listings}
\usepackage{makecell}

\usepackage{multirow}
\usepackage{multicol}
\newcommand{\logan}{CinePile\xspace}
\newcommand{\fullannot}{scene-text-annotation\xspace}
\newcommand{\gpt}{GPT-4\xspace}
\newcommand{\gpturbo}{GPT-3.5\xspace}
\newcommand{\gemini}{Gemini\xspace}
\usepackage{graphicx,pifont}

\usepackage{tcolorbox}
\definecolor{lightroyalblue}{HTML}{F6F8FD} 
\definecolor{royalblue}{HTML}{4169E1}
\definecolor{lighterblue}{HTML}{f2fafd}  
\newtcolorbox{abox}{colback=lightroyalblue,colframe=black}

\definecolor{darkblue}{RGB}{46,25, 110}
\usepackage{soul}

\hypersetup{ 
    colorlinks=true,
    linkcolor=magenta,
    filecolor=magenta,      
    urlcolor=magenta,
    citecolor=teal,
    pdftitle={CinePile A long video question answering dataset and benchmark},
    pdfpagemode=FullScreen,
    }

\newcommand{\cmark}{\textcolor{teal}{\ding{51}}}
\newcommand{\xmark}{\textcolor{red}{\ding{55}}}

\title{CinePile: A Long Video Question Answering Dataset and Benchmark}

\author{
Ruchit Rawal \textsuperscript{\rm {\color{red}\ding[1.2]{169}}} \quad
Khalid Saifullah \textsuperscript{\rm {\color{red}\ding[1.2]{169}}} \quad
Miquel Farré \textsuperscript{\rm {\color{orange}\ding[1.2]{118}}} \quad
Ronen Basri \textsuperscript{\rm {\color{black}\ding[1.2]{168}}} \quad  \\
\textbf{David Jacobs}  \textsuperscript{\rm {\color{red}\ding[1.2]{169}}} \quad
\textbf{Gowthami Somepalli}$^\star$\textsuperscript{\rm {\color{red}\ding[1.2]{169}}} \quad
\textbf{Tom Goldstein}$^\star$\textsuperscript{\rm {\color{red}\ding[1.2]{169}}}
\\
\and 
\textsuperscript{\rm {\color{red}\ding[1.2]{169}}} University of Maryland, College Park
\quad
\textsuperscript{\rm {\color{orange}\ding[1.2]{118}}} Hugging Face
\quad
\textsuperscript{\rm {\color{black}\ding[1.2]{168}}} Weizmann Institute of Science\\ \and
\url{https://hf.co/datasets/tomg-group-umd/cinepile}
}

\iclrfinalcopy 

\begin{document}

\maketitle

\begin{figure}[th]
  \centering
   \includegraphics[width=0.95\textwidth]{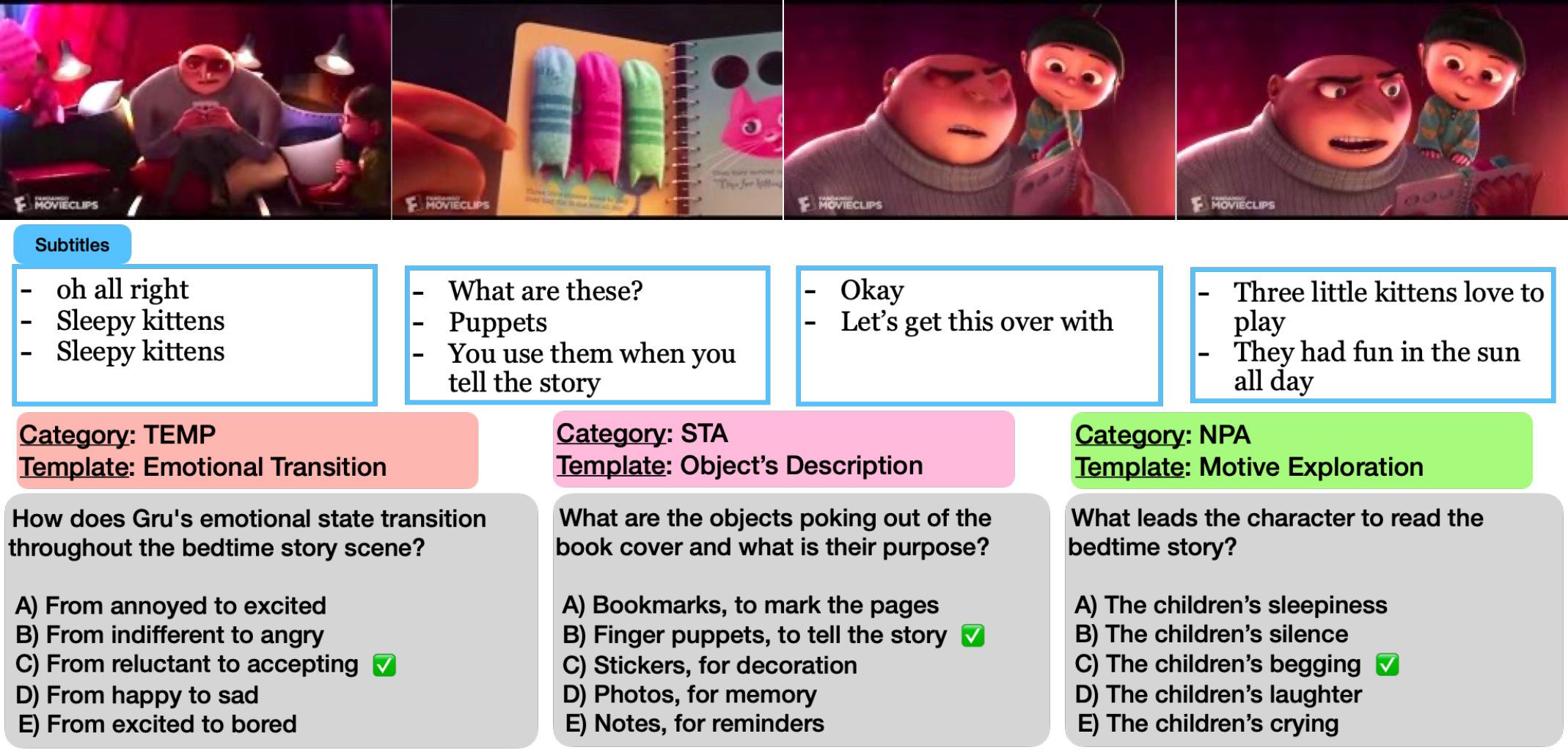}
     \caption{A sample clip (from \href{https://www.youtube.com/watch?v=Z4DDrBjEBHE&t=1s}{here}) and corresponding MCQs from CinePile.}

  \label{fig:teaser_video_qa_examples}
\end{figure}
\begin{abstract}
  Current datasets for long-form video understanding often fall short of providing genuine long-form comprehension challenges, as many tasks derived from these datasets can be successfully tackled by analyzing just one or a few random frames from a video. To address this issue, we present a novel dataset and benchmark, \logan, specifically designed for authentic long-form video understanding. This paper details our innovative approach for creating a question-answer dataset, utilizing advanced LLMs with human-in-the-loop and building upon human-generated raw data. Our comprehensive dataset comprises 305,000 multiple-choice questions (MCQs), covering various visual and multimodal aspects, including temporal comprehension, understanding human-object interactions, and reasoning about events or actions within a scene. Additionally, we fine-tuned open-source Video-LLMs on the training split and evaluated both open-source and proprietary video-centric LLMs on the test split of our dataset. The findings indicate that although current models underperform compared to humans, fine-tuning these models can lead to significant improvements in their performance.
\end{abstract}

\let\oldaddcontentsline\addcontentsline
\renewcommand{\addcontentsline}[3]{}

\section{Introduction}
\label{sec:intro}

Large multi-modal models offer the potential to analyze and understand long, complex videos.  However, training and evaluating models on video data offers difficult challenges.  Most videos contain dialogue and pixel data and complete scene understanding requires both.  Furthermore, most existing vision-language models are pre-trained primarily on still frames, while understanding long videos requires the ability to identify interactions and plot progressions in the temporal dimension. 

In this paper, we introduce \logan, a large-scale dataset consisting of $\sim$~305k question-answer pairs from 9396 videos, split into train and test sets. Our dataset emphasizes question diversity, and topics span temporal understanding, perceptual analysis, complex reasoning, and more. It also emphasizes question difficulty, with humans exceeding the best commercial vision/omni models by approximately 25\%, and exceeding open source video understanding models by 37\%.

We present a scene and a few question-answer pairs from our dataset in \cref{fig:teaser_video_qa_examples}. Consider the first question, \texttt{How does Gru's emotional state 
transition throughout the scene?} For a model to answer this correctly, it needs to understand both the visual and temporal aspects, and even reason about the plot progression of the scene. To answer the second question, \texttt{What are the objects poking out of the book cover and what is their purpose}, the model must localize an object in time and space, and use its world knowledge to reason about their purpose.

\logan addresses several weaknesses of existing video understanding datasets.  First, the large size of \logan enables it to serve as both an instruction-tuning dataset and an evaluation benchmark.  We believe the ability to do instruction tuning for video at a large scale can bridge the gap between the open-source and commercial video understanding models. Also, the question diversity in \logan makes it a more comprehensive measure of model performance than existing benchmarks.  Unlike existing datasets,  \logan does not over-emphasize on purely visual questions (e.g., \texttt{What color is the car?}), or on classification questions (e.g., \texttt{What genre is the video?}) that do not require temporal understanding.  Rather, \logan is comprehensive with diverse questions about vision, temporal, and narrative reasoning with a breakdown of question types to help developers identify blind spots in their models.  

The large size of \logan is made possible by our novel pipeline for automated question generation and verification using large language models.  Our method leverages large existing sets of audio descriptions that have been created to assist the vision impaired.  We transcribe these audio descriptions and align them with publicly available movie video clips from YouTube.  Using this detailed human description of scenes, powerful LLMs are able to create complex and difficult questions about the whole video without using explicit video input.  At test time, video-centric models must answer these questions from only the dialogue and raw video, and will not have access to the hand-written descriptions used to build the questions. We release the prompts for generating the question answers, the code for model evaluation, and the dataset splits in the Appendix.
\section{Creating a long video understanding benchmark}
\label{sec:method}

Our dataset curation process has four primary components 1) Collection of raw video and related data. 2) Generation of question templates. 3) Automated construction of the Q\&A dataset using video and templates, and 4) Application of a refinement pipeline to improve or discard malformed Q\&A pairs.

\subsection{Data collection and consolidation}
\label{subsec:data_collection}
\vspace{-0.5em}
We obtain clips from English-language films from the YouTube channel  {\em MovieClips}\footnote{\url{https://www.youtube.com/@MOVIECLIPS}}. This channel hosts self-contained clips, each encapsulating a major plot point, facilitating the creation of a dataset focused on understanding and reasoning. Next, we collected Audio Descriptions from AudioVault\footnote{\url{https://audiovault.net/movies}}.

\vspace{-0.5em}
\noindent \textbf{Getting visual descriptions of video for free.} Audio descriptions (ADs) are audio tracks for movies that feature a narrator who explains the visual elements crucial to the story during pauses in dialogue.  They have been created for many movies to assist the vision impaired.  
The key distinction between conventional video caption datasets and ADs lies in the contextual nature of the latter. In ADs, humans emphasize the important visual elements in their narrations, unlike other video caption datasets, which tend to be overly descriptive. We use the audio descriptions as a proxy for visual annotation in the videos for our dataset creation.
\smallskip

\noindent \textbf{Scene localization in AD.} 
The video clips we have gathered are typically 2-3 minutes long, while Audio Descriptions (ADs) cover entire movies. To align descriptions with video, we transcribe the audio from both the movie clip and the whole movie AD file using an Automatic Speech Recognition (ASR) system \texttt{WhisperX}~\citep{bain2022whisperx}, an enhanced version of \texttt{Whisper}~\citep{radford2023robust} designed to offer quicker inference and more precise word-level timestamps. We then embed the first 3 and last 3 lines of the text transcription of a YouTube movie clip using a sentence embedding model, \texttt{WhereIsAI/UAE-Large-V1}. We similarly embed all the sentences in the corresponding movie AD file. We then localize the YouTube clip within the AD file via the rolling window algorithm. We then extract all AD data that lies between the matched start and end of the movie clip embeddings. This localized text contains both the visual elements and the dialogue for the given YouTube clip. This serves as a base text for creating the QA dataset 
For the rest of the paper, we will refer to the human-written description of the scene as ``visual description'' and the speaking or dialogue part of the video as ``dialogue''. When combined, we will refer to both data sources as ``\textbf{\fullannot}''. 

\smallskip
\noindent \textbf{Sentence classification.}
When we transcribe an AD file, the text contains a human's visual descriptions and the movie's dialogue. However, the transcription model does not label whether a given sentence belongs to a visual description or a dialogue. Since we planned to create a few questions solely on the visual components of the video, the distinction is important to us. To categorize each sentence as either visual or dialogue, we fine-tuned a BERT-Base model~\citep{devlin2018bert} using annotations from the MAD dataset~\citep{soldan2022mad}, which contains labels indicating whether a sentence is a dialogue or a visual description. We applied a binary classification head for this task. For training the classification model, we split the MAD dataset annotations into an 80-20 training-evaluation split. The model achieves 96\% accuracy on eval split after 3 epoch training. Qualitatively, we observed that the model accurately classifies sentences in the data we curated, distinguishing effectively between dialogue and visual description content. 

\footnotetext{Icons in the figures are sourced from \href{https://www.flaticon.com/}{Flaticon}.}

\begin{figure}[tb]
  \centering
   \includegraphics[width=0.9\textwidth]{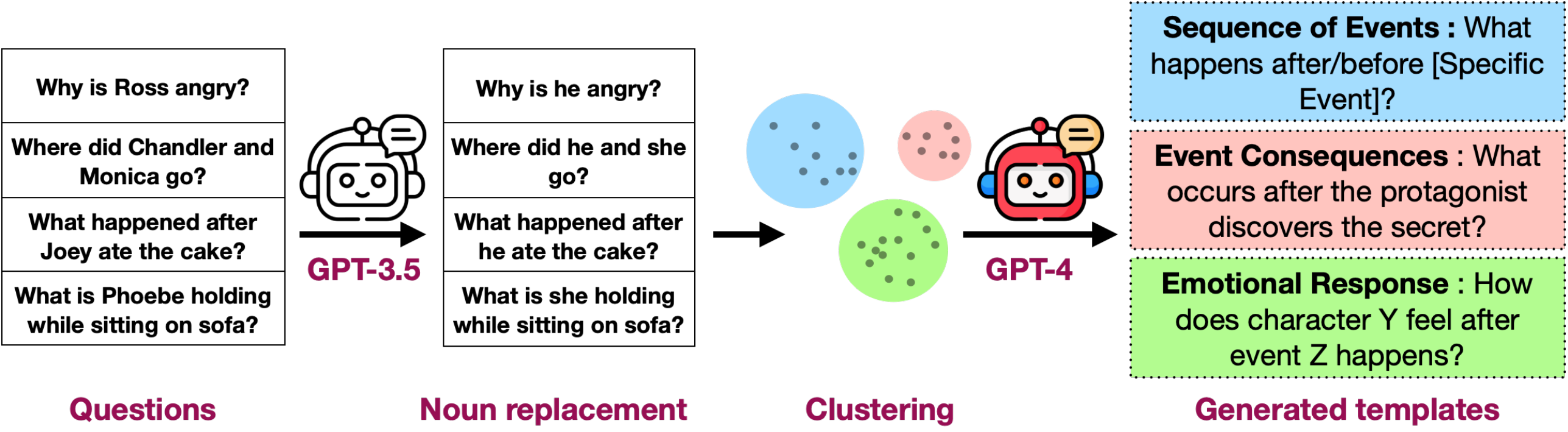}
  \caption{\textbf{Question template generation pipeline}: We begin by substituting the first names in human-written source questions and then cluster them. We then feed a selection of questions from each cluster into \gpt, which outputs ``question templates'' used in the next stage of dataset creation. See \cref{subsec:question_template_gen} for more details.
  }
  \label{fig:question_template_gen}
\end{figure}

\begin{figure}[!th]
\scriptsize
\begin{tcolorbox}[
    colback=violet!10!white, 
    colframe=violet!35!white, 
    coltitle=black, 
    title=\textbf{Question Template Automation}, 
    fonttitle=\bfseries\small, 
    arc=4mm, 
    enhanced, 
    attach boxed title to top left={yshift=-\tcboxedtitleheight/2, xshift=10pt}, 
    boxed title style={
        enhanced,
        colback=violet!10!white,
        colframe=violet!35!white,
        arc=0mm,
        left=0pt,
        right=0pt,
        boxsep=0pt
    }
]

{
Generate a few templates based on the following questions \\

\texttt{[Question 1]} Where did the couple meet? \\
\texttt{[Question 2]} What holiday is coming up when they meet? \\
\texttt{[Question 3]} What made him break with her? \\

\textbf{Response} Based on the nature of these questions, I can identify a few underlying templates that encompass most of them: \\

\texttt{[Template 1]} Catalytic Actions \\
\texttt{[Proto Question 1]} What event acts as a catalyst for the character's next major decision? \\

\texttt{[Template 2]} Setting and Context \\
\texttt{[Proto Question 2]} Where does this interaction take place, and how does the location impact the conversation? \\
...
}
\end{tcolorbox}
\caption{{\bf Extracting templates from human-generated questions.} We share 10 questions from each cluster, and prompt an LLM to create a few templates and a prototypical question. See \cref{fig:question_template_gen} and \cref{subsec:question_template_gen} for details.}
\label{fig:question_themes_gpt4_prompt}
\end{figure}
\vspace{-0.4em}

\subsection{Automated Question Templates}
\label{subsec:question_template_gen}
\vspace{-0.5em}

Many prominent video question-answering benchmarks were written by human annotators. The question-answer pairs are typically curated in one of two ways:   1) Human annotators are given complete freedom to ask questions about a given scene~\citep{tapaswi2016movieqa} 2) They are asked to focus on specific aspects and are trained or provided with examples of questions, encouraging them to write more questions in a similar style~\citep{xiao2021next,li2020hero,lei2018tvqa,patraucean2024perception}. For instance, in the Perception Test Benchmark~\citep{patraucean2024perception}, annotators are directed to concentrate on temporal or spatial aspects, while for the Next-QA dataset~\citep{xiao2021next}, annotators mainly focused on temporal and causal action reasoning questions.

\begin{figure}[tb]
  \centering
   \includegraphics[width=0.9\textwidth]{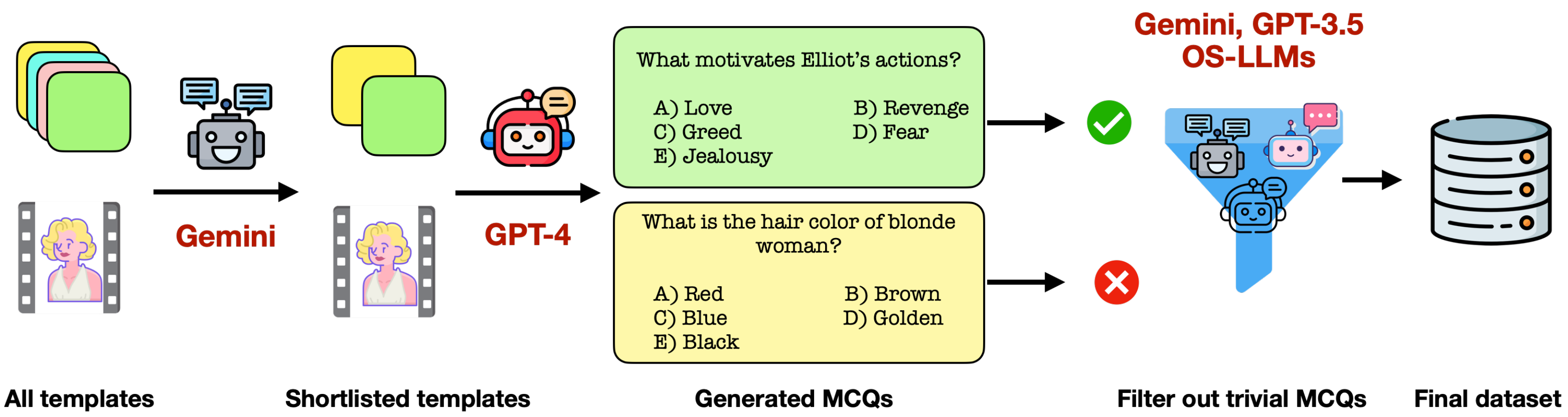}
  \caption{\textbf{Automated QA Generation and Filtering.} Begins with a set of automated templates and scenes. Filter out the templates relevant to each scene. Next, pass these templates along with the annotated-scene-text to \gpt, which is then used to create multiple-choice questions (MCQs). The generated MCQs are then subjected to numerous filters to curate the final dataset. For more detailed information, refer to \cref{subsec:auto_qa_gen} and \cref{subsec:dataset_filtering} }
  \label{fig:qa_gen_pipeline}
\end{figure}

During early experiments, we found that giving a range of templates and \fullannot to an LLM helped create more detailed, diverse, and well-formed questions. Thus, we adopted a template-based approach for question generation. Instead of limiting questions to a few hand-curated themes, we propose a pipeline to create templates from human-generated questions (shown in \cref{fig:question_template_gen}).
 
Our starting point is approximately 30,000 human-curated questions from the MovieQA~\citep{tapaswi2016movieqa}, TVQA~\citep{lei2018tvqa}, and Perception Test~\citep{patraucean2024perception} datasets. We cluster these questions, select a few representatives per cluster, and then use \gpt to discern the underlying themes and write a prompt. First, we preprocess the questions by replacing first names and entities with pronouns, as BERT~\citep{reimers2019sentence} embeddings over-index on proper nouns, hence the resultant clusters end up with shared names rather than themes. For instance, `Why is Rachel hiding in the bedroom?' is altered to `Why is she hiding in the bedroom?'. We used GPT-3.5 to do this replacement, as it handled noun replacement better than many open-source and commercial alternatives.  The modified questions are then embedded using \texttt{WhereIsAI/UAE-Large-V1}, a semantic textual similarity model which is a top performer on the MTEB leaderboard\footnote{\url{https://huggingface.co/spaces/mteb/leaderboard}}. When the first names were replaced, we observed significant repetition among questions, which prompted us to duplicate them, ultimately resulting in 17,575 unique questions. We then perform k-means clustering to categorize the questions into distinct clusters. We experimented with different values of $k =10,50,100$. Qualitatively, we found $k=50$ to be an optimal number of clusters where the clusters are diverse and at the same time clusters are not too specific. For example, we see a `high-school dance' cluster when $k=100$, and these questions are merged into an `event' cluster when we reduce $k$ to 50. The Perception Test questions are less diverse as human annotators were restricted to creating questions based on a small number of themes, so we used $k=20$ for this set. 
The number of questions in each cluster ranges from 60 to 450. We selected 10 random questions from each, and used them to prompt \gpt to create relevant question templates, as illustrated in \cref{fig:question_themes_gpt4_prompt}. We did ablations by selecting the closest 10 questions to the cluster center, however qualitatively observed that random questions produced more general/higher quality templates.

We generate four templates for each question cluster, resulting in around 300 templates across three datasets. We then manually reviewed all 300 templates, eliminating those that were overly specific and merging similar ones. Overly specific templates and their proto-questions looked like ``\textbf{Pre-wedding Dilemmas:} \texttt{What complicates character Z's plans to propose marriage to their partner?}'' and ``\textbf{Crime and Consequence:} \texttt{What is the consequence of the character's criminal actions?}''. The authors also added a many templates that were complimentary to the auto-generated ones. This process resulted in 86 unique templates.  Following that, we manually binned these into five high-level categories: Character and Relationship Dynamics, Narrative and Plot Analysis, Thematic Exploration, Temporal, and Setting and Technical Analysis. For a detailed discussion on the category definitions, examples of templates, and prototypical questions from each category, please refer to the Appendix~\ref{appendix_sec:purely_perceptual} \& \ref{appendix_sec:qgen_details}.

\subsection{Automated QA generation with LLMs}
\label{subsec:auto_qa_gen}
The pipeline for generating questions is shown in \cref{fig:qa_gen_pipeline}. While the question templates are general, they might not be relevant to all the movie clips. Hence for a given scene,  we choose a few relevant question templates by providing \gemini with the \fullannot of the scene, and asking to shortlist the 20 most relevant templates to that scene, out of which we randomly select 5-6 templates. 
We then provide a commercial language model with (i) the \fullannot, which includes both visual descriptions and dialogue, (ii) the selected question template names (e.g. `Physical Possession'), (iii) the prototypical questions for the templates (e.g. ``What is [Character Name] holding''), and (iv) a system prompt asking it to write questions about the scene.
Through rigorous experimentation, we devised a system prompt that makes the model attentive to the entire scene and is capable of generating deeper, longer-term questions as opposed to mere surface-level perceptual queries.  We observed that providing the prototypical example prevents \gpt from hallucination, and also leads to more plausible multiple-choice question (MCQ) distractors.  We also found that asking the model to provide rationale for its answer enhances the quality of the questions. Additionally, we found that including timestamps for the \fullannot augments the quality of generated temporal questions.  Through this method, we were able to generate $\approx$ 32 questions per video.

After experimenting with this pipeline, we analyzed the generated QA pairs and noticed a consistent trend: most questions are focused on reasoning or understanding. For diversity, we also wanted to include purely perceptual questions. To achieve this, we introduced additional hand-crafted prompt templates for perceptual questions and also templates for temporal questions. While \gpt performs well across all question templates, we found that \gemini excels particularly with perceptual templates. Therefore, we utilized \gemini to generate a segment of perceptual questions in the dataset, while using \gpt for reasoning templates. Our experiments with open-source models indicated subpar question quality, despite extensive prompt tuning. We present example questions and a quantitative investigation into the quality of the generations produced by \gpt and \gemini in Appendix~\ref{appendix_sec:prop_model_ques_gen}. Moreover, we provide the prompt we use question-answer generation in Appendix~\ref{appendix_sec:qa_prompt}.


\begin{figure}
     \centering
     \begin{subfigure}[b]{0.35\textwidth}
         \centering
         \includegraphics[width=\textwidth]{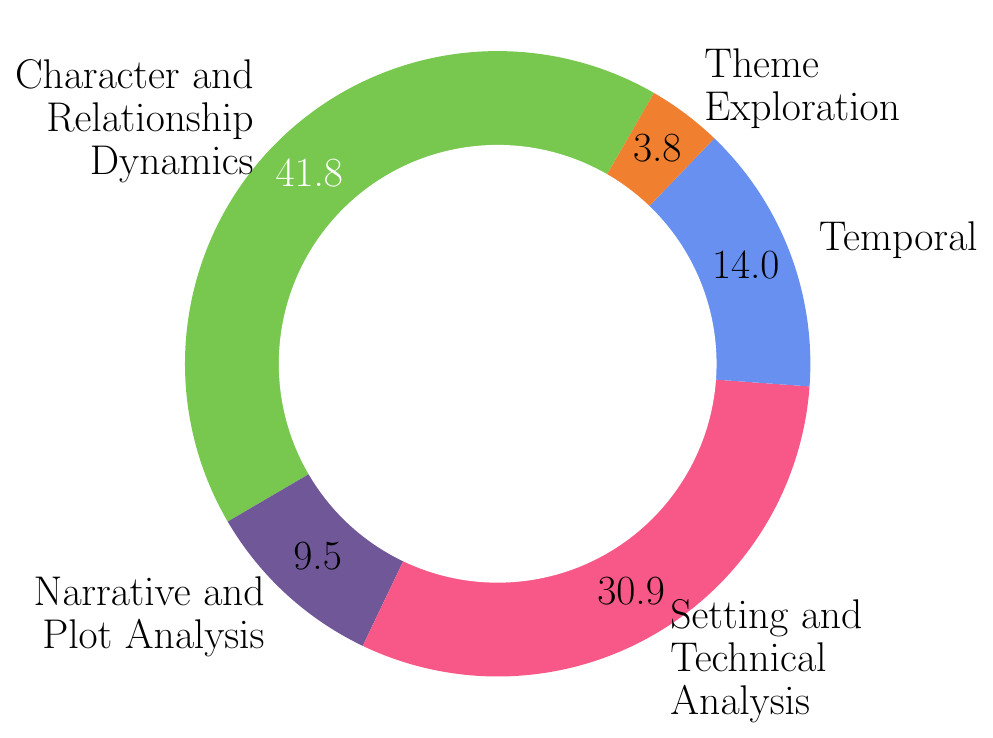}
         \label{fig:question_theme_distribution}
     \end{subfigure}
     \hfill
     \begin{subfigure}[b]{0.26\textwidth}
         \centering
         \includegraphics[width=\textwidth]{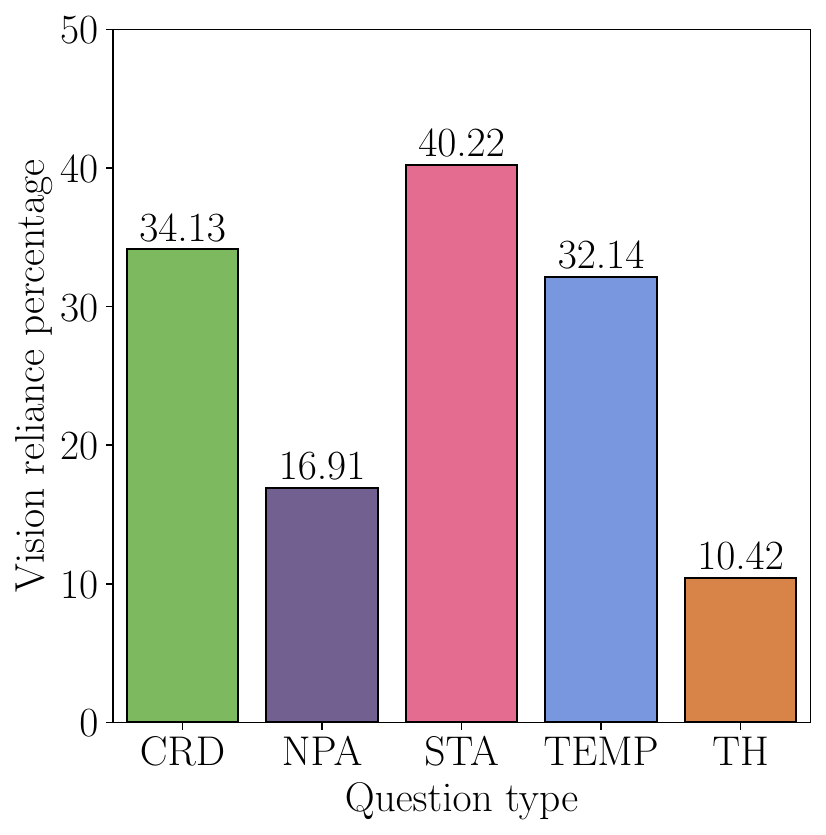}
    \label{fig:vision_reliance}
     \end{subfigure}
     \hfill
     \begin{subfigure}[b]{0.26\textwidth}
         \centering
         \includegraphics[width=\textwidth]{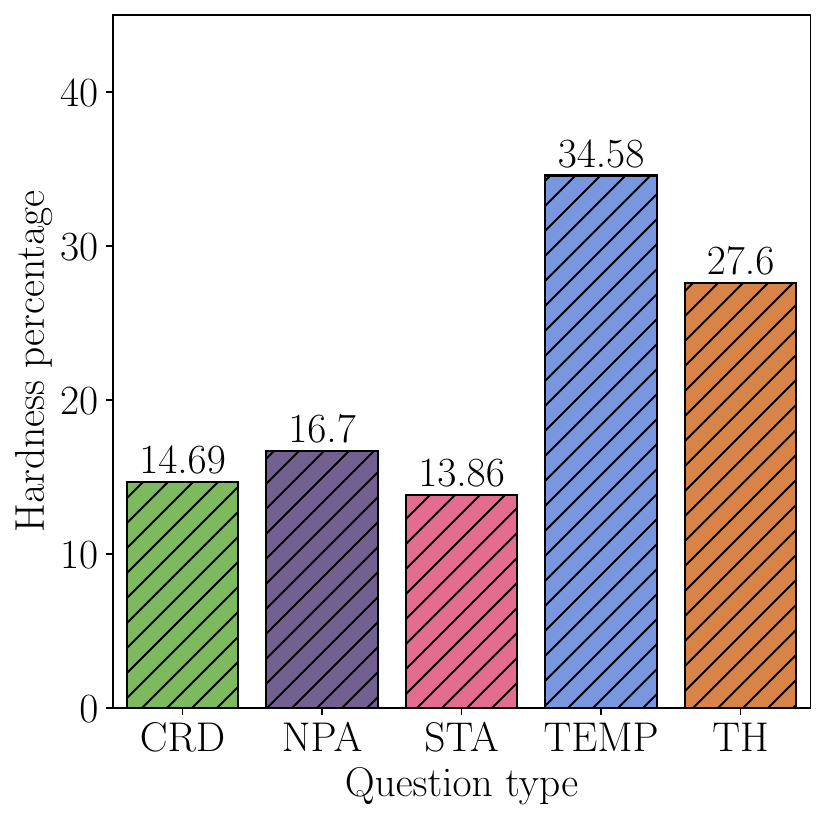}
    \label{fig:hardness_metrics}
     \end{subfigure}
        \caption{Test split statistics. \textbf{Left:} Question category composition in the dataset. \textbf{Middle:} Percentage of vision-reliant questions across categories. \textbf{Right:} Percentage of hard questions per question category type. TEMP - Temporal, CRD - Character and Relationship Dynamics, NPA - Narrative and Plot Analysis, STA - Setting and Technical Analysis, TH - Thematic Exploration.  The colors correspond to the same categories across the plots. Refer to the Appendix for corresponding plots of train split. }
       \label{fig:dataset_stats_figure}
\end{figure}
\setlength{\textfloatsep}{8pt} 
\setlength{\floatsep}{8pt} 

\vspace{-1.4em}
\subsection{Dataset quality evaluation and adversarial refinement} 
\label{subsec:dataset_filtering}
\vspace{-0.5em}
While the process above consistently produces well-formed and answerable questions, we observed that some questions are either trivial, with answers embedded within the question itself, or pertaining to basic world concepts that do not require viewing the clip. To identify these, we evaluated our dataset with the help of a few LLMs on the following axes and we improved the quality of those whenever possible. In the few instances where this was not possible, we removed the questions from the dataset or computed a metric that the users can use in the downstream tasks.

\noindent \textbf{Degeneracy and educated guessing.} A question is considered degenerate if the answer is implicit in the question itself, e.g., \texttt{What is the color of the pink house?}.  Similarly, an educated guessing  is the most probable answer to the question based on general knowledge, context, or common sense, e.g. \texttt{What is the bartender using the shaker for? a) \textbf{prepare a cocktail} b) do groceries c) collect tips }.
Based on an investigation of a subset of the dataset, we found that such questions constituted only a small fraction. However, since manually reviewing all the questions was impractical, we employed three distinct language models (LMs) to identify weak Q\&As: \gemini~\citep{team2023gemini}, \gpturbo~\citep{achiam2023gpt}, and Phi-1.5~\citep{li2023textbooks}. In order to do this,  we presented only the questions and answer choices to the models, omitting any context, and calculated the accuracy for each question across multiple models. If multiple models with different pre-training or post-training setups all correctly answer a question, it is likely that the answer was implicit, rather than due to biases of any one.

\begin{figure}
    \centering
    \includegraphics[width=\linewidth]{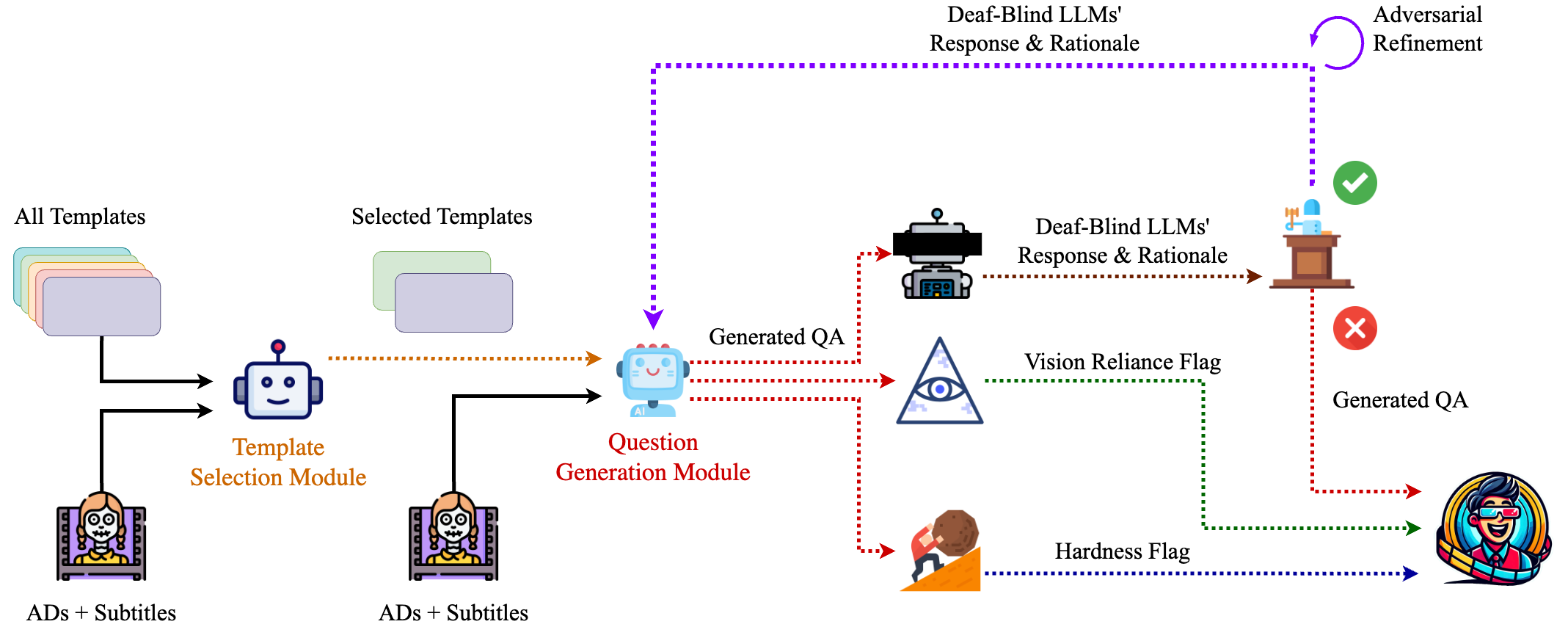}
    \caption{The pipeline for template selection, question generation, and filtration begins with selecting templates relevant to the particular scene. Next, the scene-text annotations and selected templates are used to generate question-answer pairs. Finally, the filtration step refines the pipeline by ensuring the questions are unanswerable without context, while also computing other helpful flags.}
    \label{fig:compl_pipeline}
\end{figure}

\noindent \textbf{Adversarial Refinement.} After identifying weak Q\&A pairs, we initiated an \textit{adversarial refinement} process to repair these Q\&A pairs. The goal was to modify the questions and/or answer choices so that a language model could no longer answer them correctly using only implicit clues within the question and answer choices themselves. To achieve this, we used a large language model (LLM) to identify and explain why a question could be answered without extra context. Specifically, when the LLM answered a question correctly, we asked it to provide a rationale for its choice. This rationale helped us detect hidden hints or biases in the question. We then fed this rationale into our question-generation model, instructing it to modify the question and/or answer choices to eliminate these implicit clues. This process continued in a loop until the LLM could no longer answer the question correctly (after adjusting for chance performance), with a maximum of five attempts per question. 
Given the repetitive and computationally intensive nature of this process, we required a powerful yet accessible LLM that could run locally, avoiding issues with API limits, delays, and costs associated with cloud-based services. As a result, we selected LLaMA 3.1 70B \citep{dubey2024llama}, an open-source model that met these desiderata. 
Through this adversarial refinement process, we successfully corrected approximately 90.94\% of the weak Q\&A pairs in the training set and 90.24\% of the weak Q\&A pairs in the test set. Finally, we excluded the unfixable Q\&A pairs  from the evaluation split ($\sim$ 80 Q\&A) of our dataset but retained them in the training set ($\sim$ 4500 Q\&A).

\noindent \textbf{Vision Reliance.} When generating the multiple-choice questions (MCQs), we considered the entire scene without differentiating between visual text and dialogue. Consequently, some questions in the dataset might be answerable solely based on dialogue, without the necessity of the video component. For this analysis, we utilized the \gemini model. The model was provided with only the dialogue, excluding any visual descriptions, to assess its performance. If the model correctly answers a question, it is assigned a score of 0 for the visual dependence metric; if it fails, the score is set at 1. In later sections, we present the distribution of the visual dependence scores across different MCQ categories.

\noindent \textbf{Hardness.} Hardness refers to the inability to answer questions, even when provided with full context used to create the questions in the first place (i.e., the subtitles \& visual descriptions). For this purpose, we selected the \gemini model, given its status as one of the larger and more capable models. Unlike accuracy evaluation, which uses only video frames and dialogues (subtitles), the hardness metric includes visual descriptions as part of the context given to the model. After this, the authors reviewed all the questions flagged as ``hard" for verification and fixed any minor issues, if present.

In addition, the authors went through the question in the evaluation split across multiple iterations, and fixed any systemic errors that arose in the pipeline. Furthermore, we conducted a human study to identify potential weaknesses, and we discuss our findings in \cref{appendix:human_study_details}.

\vspace{-1em}
\section{A look at the dataset}
\label{sec:dataset_details}

\begin{table}[]
\centering 
\caption{We compare our dataset, \logan against the pre-existing video-QA datasets. Our dataset is both large and diverse. Multimodal refers to whether both the video and audio data is used for question creation and answering. For understanding different QA types, refer to \cref{subsec:auto_qa_gen}}
\label{table:dataset_comparison_wothers}
\scalebox{0.6}{
\begin{tabular}{llllp{4em}p{2em}p{2em}p{2em}p{2em}}
\toprule
\multirow{2}{*}{\textbf{Dataset}} & \multirow{2}{*}{\textbf{Annotation}} & \multirow{2}{*}{\textbf{Num QA}} & \multirow{2}{*}{\textbf{Avg sec}} & \multirow{2}{*}{\textbf{Multimodal}} & \multicolumn{4}{c}{\textbf{QA Type}}                  \\ 
                         &                             &                         &                          &                             & Temporal & Attribute & Narrative & Theme \\ \midrule
TGIF-QA~\citep{jang2017tgif}                & Auto                        & 165,165                 & 3                        & \xmark                       & \cmark  & \xmark        & \xmark    & \xmark     \\
MSRVTT-QA~\citep{xu2017video}                & Auto                        & 243,690                 & 15                       & \xmark                       & \xmark  & \cmark        & \xmark    & \xmark     \\
How2QA~\citep{li2020hero}                   & Human                       & 44,007                  & 60                       & \xmark                       & \cmark  & \cmark        & \xmark    & \xmark     \\
NExT-QA~\citep{xiao2021next}                  & Human                       & 52,044                  & 44                       & \xmark                       & \cmark  & \cmark        & \xmark    & \xmark     \\
EgoSchema~\citep{mangalam2024egoschema}                & Auto                        & 5,000                   & 180                      & \xmark                       & \cmark  & \cmark        & \cmark    & \xmark     \\
MovieQA~\citep{tapaswi2016movieqa}                  & Human                       & 6,462                   & 203                      & \cmark                       & \cmark  & \cmark        & \cmark    & \xmark     \\
TVQA~\citep{lei2018tvqa}                     & Human                       & 152,545                 & 76                       & \cmark                       & \cmark  & \cmark        & \cmark    & \xmark     \\
Perception Test~\citep{patraucean2024perception}  & Human   & 44,000                  & 23   & \cmark                       & \cmark  & \cmark        & \xmark    & \xmark    \\
MoVQA~\citep{zhang2023movqa}  & Human  & 21,953                  & 992                      & \cmark                       & \cmark  & \cmark        & \cmark    & \xmark    \\
IntentQA \citep{li2023intentqa} & Human & 16,297 & Unkown  & \cmark & \cmark & \xmark & \xmark & \xmark \\
Video-MME \citep{fu2024video}  & Human & 2,700 & 1017.9 & \cmark                       & \cmark  & \cmark        & \cmark    & \xmark \\
MVBench \citep{li2024mvbench} & Auto & 4,000 & 16 & \cmark                       & \cmark  & \cmark        & \xmark    & \xmark \\
Video-Bench \citep{ning2023video} & Human + Auto & 17,036 & 56 & \cmark                       & \cmark  & \cmark        & \xmark    & \xmark \\
LVBench \citep{wang2024lvbench} & Human & 1,549 & 4,101 &\cmark   & \cmark  & \cmark        & \cmark    & \xmark \\
\midrule
\textbf{\logan (Ours)} & Human + Auto  &    303,828  & 160  & \cmark                       & \cmark  & \cmark        & \cmark    & \cmark    \\
\bottomrule
\end{tabular}
}
\end{table}

\vspace{-0.5em}
In the initial phase of our dataset collection, we collected $\sim$15,000 movie clips from channels like MovieClips on YouTube. We filtered out clips that did not have corresponding recordings from Audiovault, as our question generation methodology relies on the integration of visual and auditory cues—interleaved dialogues and descriptive audio—to construct meaningful questions. We also excluded clips with low alignment scores when comparing the YouTube clip's transcription with the localized scene's transcription in the Audio Description (AD) file as discussed in \cref{subsec:data_collection}. This process resulted in a refined dataset of 9396 movie clips. The \textbf{average video length in our dataset is $\sim$160 sec}, significantly longer than many other VideoQA datasets and benchmarks.
We split 9396 videos into train and test splits of 9248 and 148 videos each. We made sure both the splits and the sampling preserved the dataset's diversity in terms of movie genres and release years. We follow the question-answer generation and filtering pipeline which was thoroughly outlined in \cref{sec:method}. We ended up with \textbf{298,887 training points and 4,941 test-set points} with around 32 questions per video scene. Each MCQ contains a question, answer, and four distractors. As a post hoc step, we randomized the position of the correct answer among the distractors for every question, thus eliminating any positional bias. We filtered out the degenerate questions from the test split, however, we left them in the train set, since those questions are harmless and might even teach smaller models some helpful biases the larger multimodal models like \gemini might inherently possess. 

Our dataset's diversity stems from the wide variety of movie clips and different prompting strategies for generating diverse question types. Each strategy zeroes in on particular aspects of the movie content. We present a scene and example MCQs from different question templates in \cref{fig:teaser_video_qa_examples}, and many more in the Appendix.
In \cref{fig:dataset_stats_figure} (Left), we provide a visual breakdown of the various question categories in our dataset. A significant portion of the questions falls under ``Character Relationship Dynamics''. This is attributed to the fact that a large number of our automated question templates, which were derived from human-written questions belonged to this category. This is followed by ``Setting and Technical Analysis'' questions, which predominantly require visual interpretation.
We display the metrics for vision reliance and question hardness, as discussed in \cref{subsec:dataset_filtering}, at the category level in \cref{fig:dataset_stats_figure} (Middle, Right). As anticipated, questions in the ``Setting and Technical Analysis'' category exhibit the highest dependency on visual elements, followed by those in ``Character Relationship Dynamics'', and ``Temporal'' categories. In terms of the hardness metric, the ``Temporal'' category contains the most challenging questions, with ``Thematic Exploration'' following closely behind.
Finally, we compare our dataset with other existing datasets in this field in \cref{table:dataset_comparison_wothers}, showing its superiority in both the number of questions and average video length compared to its counterparts.
\section{Model evaluation}
\label{sec:eval_sec}

\begin{table}[htp]
\centering
\caption{\textbf{Model Evaluations.} We present the accuracy of various video LLMs on the \logan's test split. We also present Human performance for comparison. We ablate the accuracies across the question categories: TEMP - Temporal, CRD - Character and Relationship Dynamics, NPA - Narrative and Plot Analysis, STA - Setting and Technical Analysis, TH - Thematic Exploration.
}
\label{tab:vlm_eval}
\renewcommand{\arraystretch}{1.2}
\scalebox{0.85}{
\begin{tabular}{@{}l|c|c|ccccc@{}}
\toprule
\textbf{Model} & \textbf{Params.} &\textbf{Avg} & \textbf{CRD} & \textbf{NPA} & \textbf{STA} & \textbf{TEMP} & \textbf{TH} \\
\midrule
Human & - & 73.21 & 82.92 & 75.00 & \textbf{73.00} & 75.52 & 64.93 \\
Human (authors) & - & \textbf{86.00} & \textbf{92.00} & \textbf{87.5} & 71.20 & \textbf{100} & \textbf{75.00} \\
\midrule
Gemini 1.5 Pro--001 & - & 60.12 & 63.90 & 70.44 & 57.85 & 46.74 & 59.87  \\
Gemini 1.5 Flash--001 & - & 58.75 & 62.82 & 69.76 & 55.99 & 44.04 & 62.67  \\
GPT-4o & - & 56.06 & 60.93 & 69.33 & 49.48 & 45.78 & 61.05  \\
GPT-4 Vision & - & 55.35 & 60.20 & 68.47 & 48.63 & 45.78 & 59.47  \\
LLaVA-OV & 7B & 49.34 & 52.13 & 59.83 & 46.54 & 37.65 & 58.42  \\
LLaVA-OV Chat & 7B & 49.28 & 52.47 & 58.32 & 46.28 & 37.79 & 58.42  \\
MiniCPM-V 2.6 & 8B & 46.91 & 50.10 & 54.21 & 44.52 & 35.61 & 54.74  \\
Claude 3 Opus & - & 45.60 & 48.89 & 57.88 & 40.73 & 37.65 & 47.89\\
VideoLLaMA2 & 7B & 44.57 & 47.44 & 54.64 & 41.91 & 34.30 & 47.37  \\
InternVL2 & 26B & 43.86 & 47.10 & 56.16 & 39.03 & 34.16 & 52.63  \\
LongVA DPO & 7B & 42.78 & 45.84 & 54.21 & 39.16 & 33.43 & 44.74  \\
InternVL-V1.5 & 25.5B & 41.69 & 45.07 & 51.19 & 38.97 & 30.09 & 45.79  \\
LongVA & 7B & 41.04 & 43.28 & 51.84 & 38.45 & 33.58 & 38.42  \\
InternVL2 & 4B & 39.89 & 42.99 & 47.73 & 36.23 & 32.99 & 41.58  \\
mPLUG-Owl3 & 8B & 38.27 & 40.91 & 45.71 & 33.86 & 33.09 & 46.20  \\
LLaVA-OV & 0.5B & 33.82 & 35.88 & 39.96 & 31.66 & 27.03 & 38.42  \\
InternVL2 & 8B & 32.28 & 35.25 & 40.39 & 28.46 & 24.71 & 38.42  \\
InternVL2 & 2B & 30.34 & 31.91 & 33.26 & 30.35 & 23.26 & 31.58  \\
VideoChat2  & 7B & 29.27 & 31.04 & 34.56 & 25.26 & 27.91 & 34.21  \\
Video LLaVa  & 7B & 25.72 & 26.64 & 32.61 & 23.63 & 23.26 & 24.74  \\
CogVLM2 & 19B & 17.16 & 18.33 & 17.06 & 17.23 & 13.08 & 18.95  \\
InternVL2 & 1B & 15.97 & 17.65 & 19.22 & 13.25 & 12.94 & 22.63  \\
Video-ChatGPT & 7B & 15.08 & 17.06 & 16.34 & 15.17 & 7.26 & 18.58  \\
mPLUG-Owl & 7.2B & 13.93 & 16.15 & 13.16 & 13.03 & 10.48 & 11.54  \\
\bottomrule
\end{tabular}
}
\end{table}

In this section, we discuss the evaluations of various closed and open-source video LLMs on our dataset, some challenges, and model performance trends. Given that our dataset consists of multiple-choice question answers (MCQs), we assess a model’s performance by its ability to accurately select the correct answer from a set of options containing one correct answer and four distractors. A key challenge in this process is reliably parsing the model’s response to extract its chosen answer and map it to one of the predefined choices. Model responses may vary in format, including additional markers or a combination of the option letter and corresponding text. Such variations necessitate a robust post-processing step to accurately extract and match the model's response to the correct option. To address these variations, we employ a two-stage evaluation method. First, a normalization function parses the model's response, extracting the option letter (A-E) and any accompanying text. This handles various formats, ensuring accurate identification. The second stage involves comparing the normalized response with the answer key, checking for both the option letter and text. If both match, a score of one is awarded; However, if only the option letter or text appears, the comparison is limited to the relevant part, and the score is assigned accordingly.

We evaluate $24$ commercial and open-source LLM models and we present their performance in \cref{tab:vlm_eval}. We discuss additional details about the evaluation timelines, model checkpoints, and compute budget in \cref{appendix_sec:addn_eval_details}.
We also present human numbers (author and non-author) for comparison. This distinction is important because the authors carefully watched the video (go back and rewatch the video if necessary) while answering the questions. This removes the carelessness errors from the human study. While commercial VLMs perform reasonably well, the very best of OSS models lag $\sim$10\% behind the proprietary models. We present a few QA's which humans got wrong and \gpt got wrong and the plausible reason for errors in ~\cref{appendix:human_study_details}.

\textbf{Gemini 1.5 Pro leads overall; LLaVA-OV tops open-source models.}
Among the various commercial VLMs analyzed Gemini 1.5 Pro performs the best, and particularly outperforms the {\gpt} models in the ``Setting and Technical Analysis'' category that is dominated by visually reliant questions focusing on the environmental and surroundings of a movie scene, and its impact on the characters.
On the contrary, we note that {\gpt} models offer competitive performance on question categories such as  ``Narrative and Plot Analysis” that revolve around the core storylines, and interaction between the key characters.
It's important to note that Gemini 1.5 Pro is designed to handle long multimodal contexts natively, while {\gpt}o and {\gpt}V don't yet accept video as input via their APIs. Therefore, we sample 10 frames per video while evaluating them. 
Gemini 1.5 Flash, a newly released lighter version of Gemini 1.5 Pro, also performs competitively, achieving 58.75\% overall accuracy and ranking second in performance. Its competitive edge over the GPT models is owing to the ``Setting and Technical Analysis" category, where it performs significantly better.
In open-source models, LLaVA-OV (One Vision) ranks as the best, achieving an overall accuracy of 49.34\%. More broadly, while the accuracy of open-source models ranges from 49.34\% to 13.93\%, it’s clear that recent models like LLaVA-OV (released August 2024), MiniCPM-V-2.6 (released August 2024), and VideoLLaMa2 (released June 2024) offer competitive performance compared to proprietary models.

\begin{figure}[ht]
    \centering
    \begin{subfigure}[b]{0.44\textwidth}
        \centering
        \includegraphics[height=4cm]{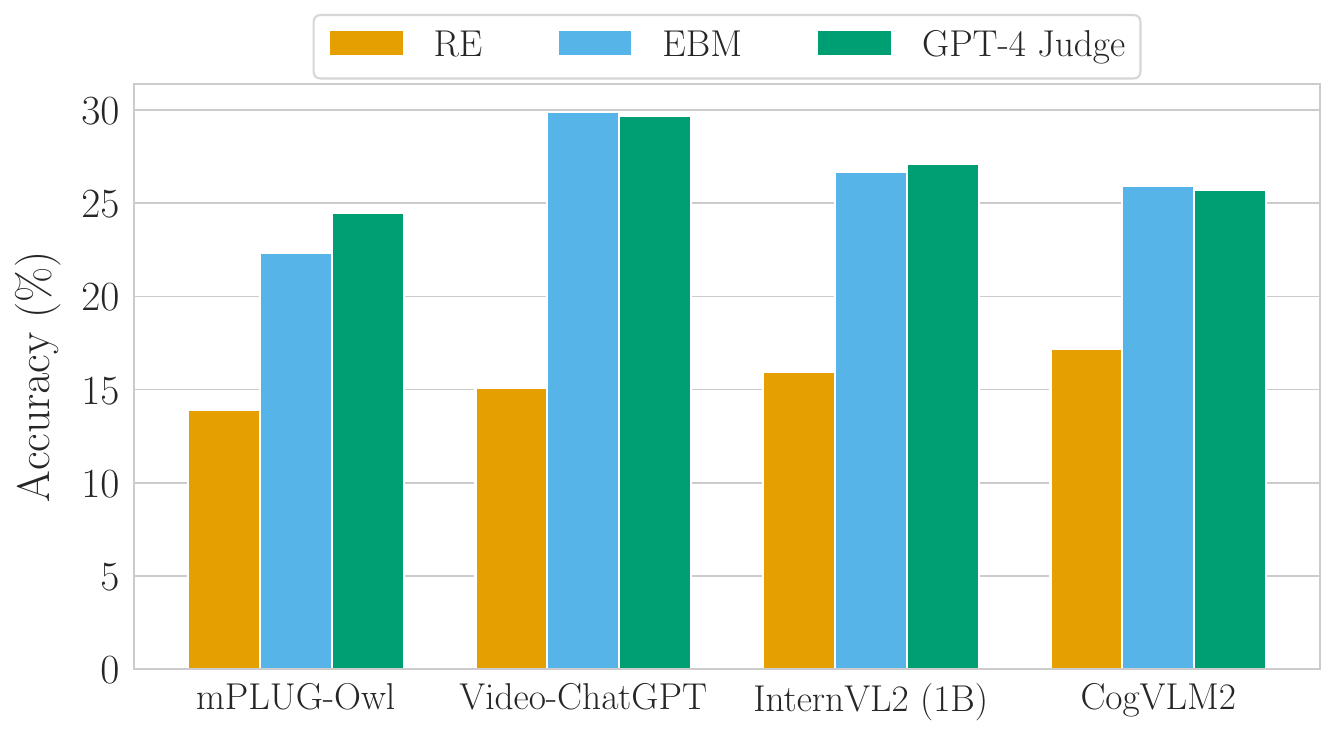}
        \caption{Different strategies for evaluating performance on \logan include: RE (Response Extraction), EBM (Embedding-Based Matching), and GPT-4 Judge (using GPT-4 to assess the raw response).}
        \label{fig:alter_strats}
    \end{subfigure}
    \hfill
    \begin{subfigure}[b]{0.48\textwidth}
        \centering
        \includegraphics[height=4cm]{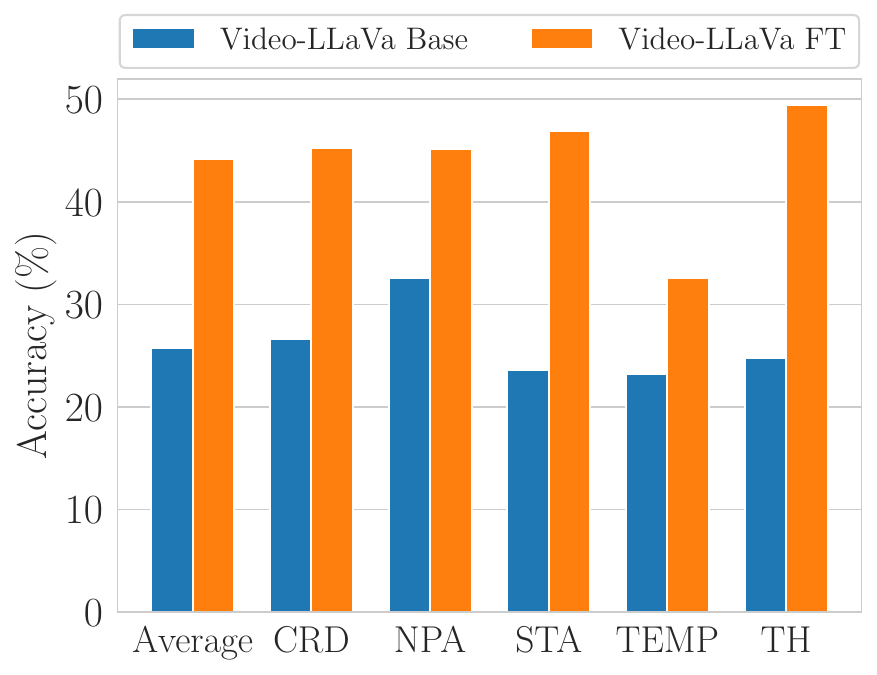}
        \caption{Comparing the performance of Video-LLaVa after fine-tuning on \logan's training set. `Average' refers to the aggregate performance, while the remaining labels represent specific question types.}
        \label{fig:ft_performance}
    \end{subfigure}
    \caption{}
\end{figure}
\setlength{\textfloatsep}{5pt} 
\setlength{\floatsep}{5pt} 

\begin{figure}[tb]
  \centering
   \includegraphics[width=\textwidth]{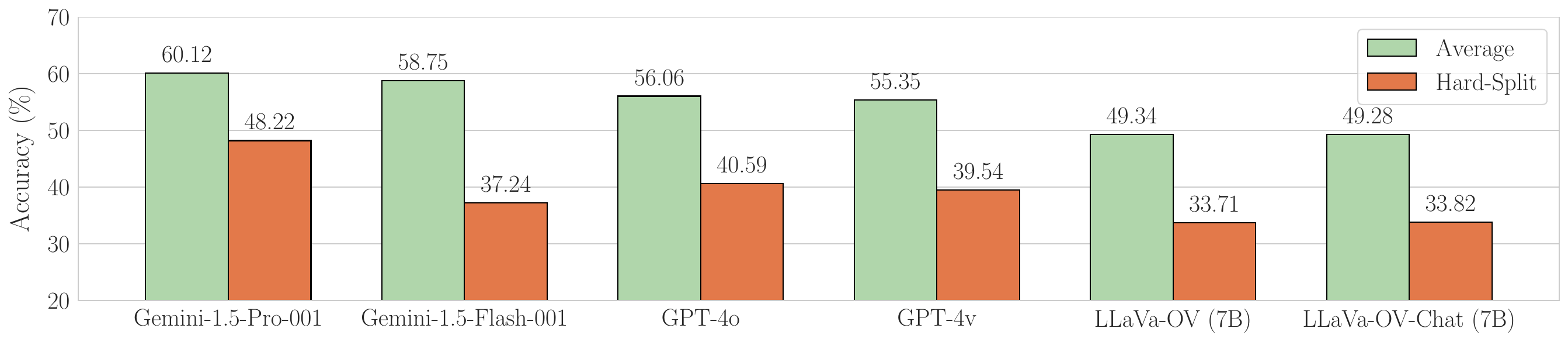}
  \caption{Models' performance on \logan test split, all questions vs hard questions.}
  \label{fig:hard_v_avg}
\end{figure}

\textbf{Performance significantly drops on the ``hard-split”.}
Additionally, as discussed in \cref{subsec:dataset_filtering}, we provide a ``hard split” in the test set consisting of particularly challenging questions. In \cref{fig:hard_v_avg}, we compare the performance of the top 6 models (in terms of average accuracy) on both the average and the hard splits of our dataset. We note that while most models suffer a performance decline of 15\%-20\% on the hard split; however, the relative ranking among the models remains unchanged.
Interestingly, Gemini 1.5 Flash suffers a decline of $\approx 21\%$ compared to $13\%$ for Gemini 1.5 Pro, underscoring the particularly severe trade-offs involved in optimizing the models for lightweight performance on more challenging samples.

\textbf{Why are (some) OSS models so far behind?}
We conducted further analyses to understand the poor performance of some open-source models, focusing on qualitative evaluations of their raw responses (Appendix~\ref{appendix_sec:additional_metrics}). Our findings indicate that a primary issue is their inability to follow instructions, often generating irrelevant or repetitive content, which hinders accurate extraction of the intended answer.
To quantify these deviations, we introduced two alternative strategies for computing accuracy: a) Embedding Similarity Matching: We compute the similarity between the model’s raw response and the various answer options within the embedding space of a sentence transformer \citep{zhang2019bertscore}. The most similar option is selected as the predicted answer. b) GPT-4 as a  judge: We use GPT-4 \citep{zheng2023judging} as an evaluator to extract the predicted answer key from the model’s raw response.
The results from these strategies are illustrated in Figure~\ref{fig:alter_strats}.
We observe that although these alternative evaluation strategies yield an improvement in the models' performance, their accuracy still falls significantly short compared to the best-performing open-source models. This suggests that the underperformance cannot be solely attributed to an inability to follow instructions. Rather, these models also exhibit fundamental limitations in video understanding capabilities. Notably, the two alternative evaluation strategies—embedding similarity matching and the use of GPT-4 as a judge—are highly consistent with each other, as well as largely aligning with the rankings obtained from the original response extraction strategy.
We provide further details and additional results based on traditional video-caption evaluation metrics, such as BertScore \citep{zhang2019bertscore}, CIDEr \citep{vedantam2015cider}, and ROUGE-L \citep{lin2004rouge}, in Appendix~\ref{appendix_sec:additional_metrics}.

\textbf{\logan's train-split helps improve performance}
In this section, we investigate the impact of \logan’s training split in enhancing the performance of open-source video LLMs. We selected Video-LLaVa as the baseline and fine-tuned it using \logan’s training data. For efficient training, we load the model using 4-bit quantization. During fine-tuning, we freeze the base model, and conduct training using Low-Rank Adaptation (LoRA) \citep{hu2021lora}. We fine-tuned the model for 5 epochs using the AdamW optimizer 
\citep{Loshchilov2017DecoupledWD}. We compare the performance of the fine-tuned Video-LLaVa against the base model, as shown in \ref{fig:ft_performance}. Our results indicate that fine-tuning led to an approximate 71\% improvement in performance (increasing accuracy from 25.72\% to 44.16\%), with gains observed consistently across all question subcategories. These results demonstrate the significant utility of CinePile’s training split in enhancing model performance. 

\textbf{Additional Ablations.} We report additional results on the effect of removing video frames on model performance in Appendix~\ref{appendix_sec:frame_rate_ablation}, performance on hard-split (for all models) in Appendix~\ref{appendix_sec:hard_split_perf_cat}.

\section{Related Work}
\label{sec:related-work}
\vspace{-0.7em}

LVU~\citep{lvu2021}, despite being one of the early datasets proposed for long video understanding, barely addresses the problem of video understanding as the main tasks addressed in this dataset are year, genre classification or predicting the like ratio for the video. A single frame might suffice to answer the questions and these tasks cannot be considered quite as ``understanding'' tasks.  MovieQA~\citep{tapaswi2016movieqa} is one of the first attempts to create a truly understanding QA dataset, where the questions are based on entire plot the movie but not localized to a single scene. On closer examination, very few questions are vision focused and most of them can be answered just based on dialogue. EgoSchema~\citep{mangalam2024egoschema} is one of the recent benchmarks, focused on video understanding which requires processing long enough segments in the video to be able to answer the questions. However, the videos are based on egocentric videos and hence the questions mostly require perceptual knowledge, rather than multimodal reasoning. Another recent benchmark, Perception Test~\citep{patraucean2024perception}, focuses on core perception skills, such as memory and abstraction, across various reasoning abilities (e.g., descriptive, predictive, etc) for short-form videos.
The MAD dataset introduced in~\citep{soldan2022mad} and expanded in \citep{han2023autoad} contains dialogue and visual descriptions for full-length movies and is typically used in scene captioning tasks rather than understanding. Another issue is this dataset does not provide raw visual data, they share only \texttt{[CLS]} token embeddings, which makes it hard to use. TVQA~\citep{lei2018tvqa} is QA dataset based on short 1-min clips from famous TV shows. The annotators are instructed to ask What/How/Why sort of questions combining two or more events in the video. MoVQA~\citep{zhang2023movqa} manually curates questions across levels multiple levels—single scene, multiple scenes, full movie— by guiding annotators to develop queries in predefined categories like Information Processing, Temporal Perception, etc. CMD~\citep{bain2020condensed} proposes a text-to-video retrieval benchmark while VCR~\citep{zellers2019recognition} introduces a commonsense reasoning benchmark on images taken from movies.

Long video understanding datasets, such as EpicKitchens~\citep{Damen2018EPICKITCHENS}, tend to concentrate heavily on tasks related to the memory of visual representations, rather than on reasoning skills.
More recently, multiple benchmarks focusing on long video understanding have been released, such as Video-MME \citep{fu2024video}, MVBench \citep{li2024mvbench}, and LVBench \citep{wang2024lvbench}. Most of these datasets require significant human effort to generate questions, with costs increasing as you move toward longer video regimes. Hence, most of them range on a scale of a few thousand question-answer pairs (while \logan ranges 70-75 $\times$ more).
We discuss works utilizing synthetic data for dataset creation in \cref{addn_rel_work}.

\logan differs from all the above datasets, having longer videos and many questions to capture the perceptual, temporal, and reasoning aspects of a video. And it is truly multimodal where the person has to watch the video as well as dialogues to answer many questions. Unlike the previous datasets with fixed templates, we automated this process on previously human-generated questions, this let us capture many more question categories compared to previous works. Lastly, our approach to dataset generation is scalable, allowing us to fine-tune video models to improve performance. Moreover, \logan can easily be extended in the future with additional videos, question categories, and more.
\section{Discussion and Conclusion}
\label{diss_conc}
\vspace{-0.7em}
In this paper, we introduced \logan, a unique long video understanding dataset and benchmark, featuring $\sim$~300k questions in the training set and $\sim$~5000 in the test split. We detailed a novel method for curating and filtering this dataset, which is both scalable and cost-effective. Additionally, we benchmarked various recent commercial video-centric LLMs and conducted a human study to gauge the achievable performance on this dataset. To our knowledge, \logan is the only large-scale dataset that focuses on multi-modal understanding, as opposed to the purely visual reasoning addressed in previous datasets. Our fine-tuning experiments demonstrate the quality of our training split. Additionally, we plan to set up a leaderboard for the test set, providing a platform for new video LLMs to assess and benchmark their performance on \logan. 

Despite its strengths, there are still a few areas for improvement in our dataset, such as the incorporation of character grounding in time. While we believe our dataset's quality is comparable to or even better than that of a Mechanical Turk annotator, we acknowledge that a motivated human, given sufficient time, can create more challenging questions than those currently generated by an LLM. Our goal is to narrow this gap in future iterations of \logan.

\bibliography{iclr2025_conference}
\bibliographystyle{iclr2025_conference}

\newpage
\appendix
\centering
\let\addcontentsline\oldaddcontentsline

{\large \textbf{CinePile: A Long Video Question Answering Dataset and Benchmark}} \\
\vspace{0.5em}{\large Appendix} \\
\vspace{1.5em}
\raggedright

\vspace{5em}
\tableofcontents

\newpage

\section{Additional movie clip \& questions examples}
\label{appendix_sec:question_examples}

\begin{figure}[!h]
    \centering
    \begin{subfigure}[b]{\textwidth}
        \includegraphics[width=\textwidth]{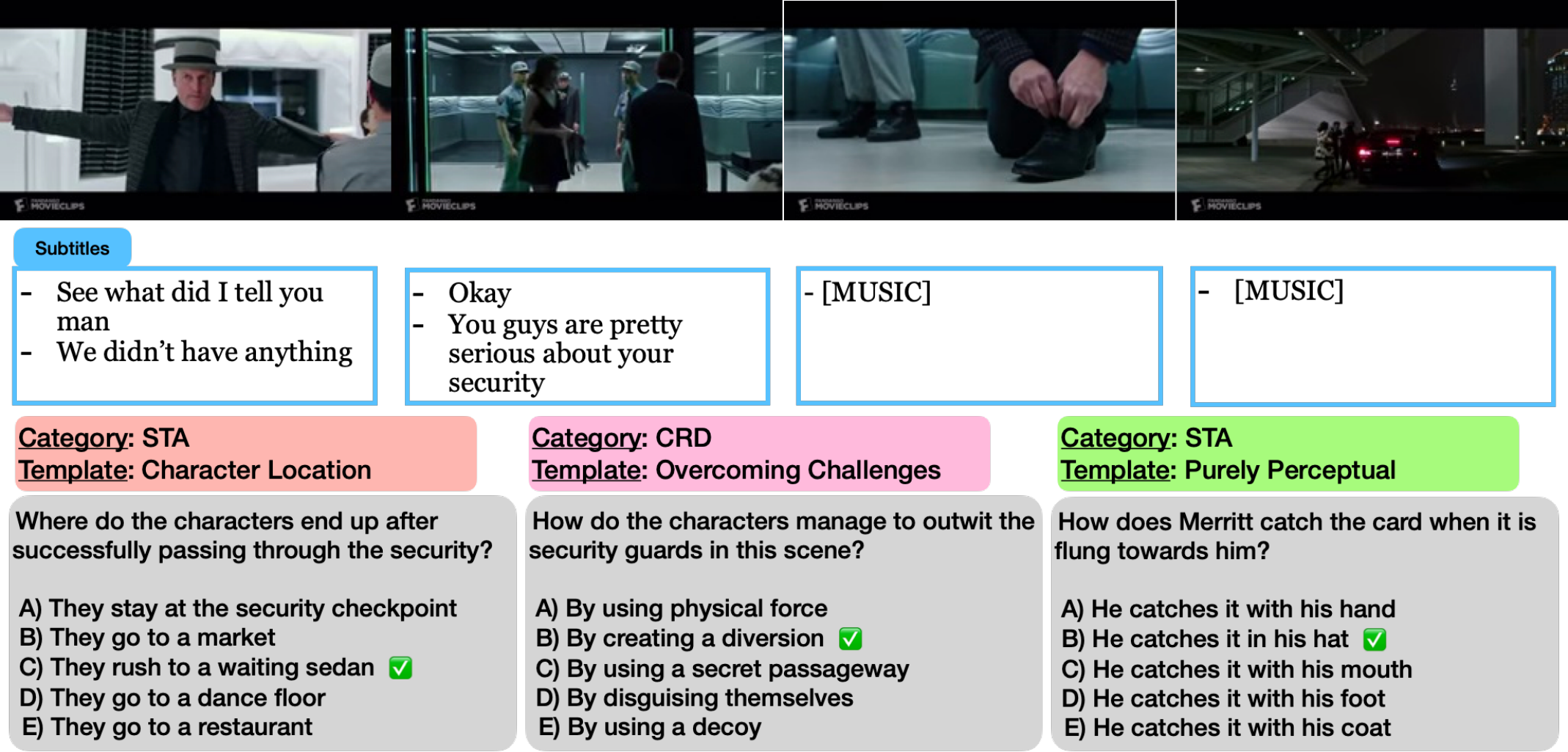}
        \caption{}
        \label{fig:scene_fig_1}
    \end{subfigure}
    
    \vspace{0.5cm}  
    
    \begin{subfigure}[b]{\textwidth}
        \includegraphics[width=\textwidth]{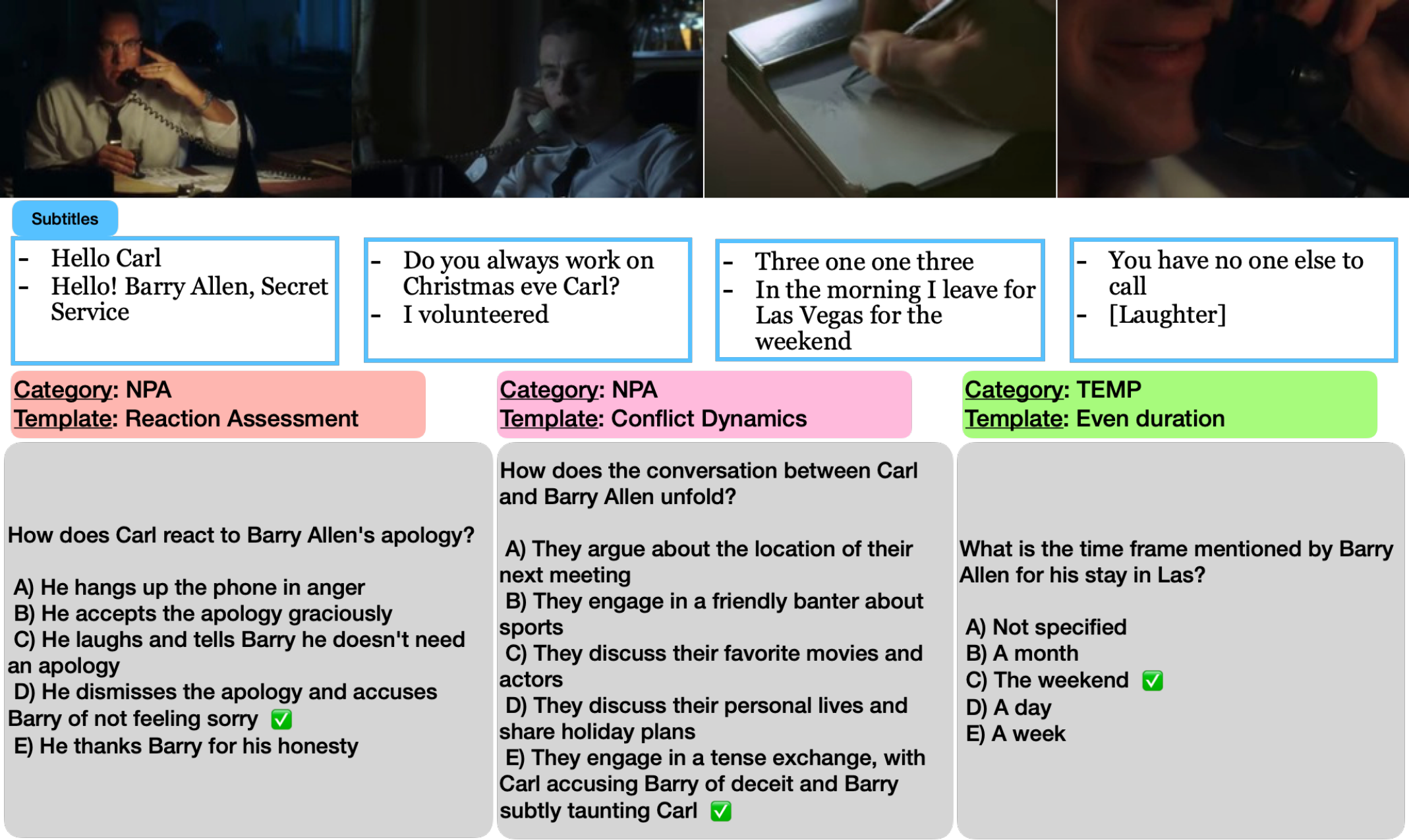}
        \caption{}
        \label{fig:scene_fig_2}
    \end{subfigure}
    
    \caption{\textbf{Example movie clip and multiple-choice questions from \logan}. The first and second rows depict a selection of image frames extracted from movie clips from (a) \href{https://www.youtube.com/watch?v=3IVugy6dK3E}{Now You See Me 2}, and (b) \href{https://youtube.com/watch?v=Zb8exHKOaK0}{Catch Me if You Can}, accompanied by their corresponding subtitles. The next row showcases example questions along with the question template shown in colored headers. TEMP refers to Temporal. Please refer to \cref{tab:question_themes} for other category acronyms.}
\end{figure}

\begin{figure}[htbp]
    \centering
    \begin{subfigure}[b]{\textwidth}
        \includegraphics[width=\textwidth]{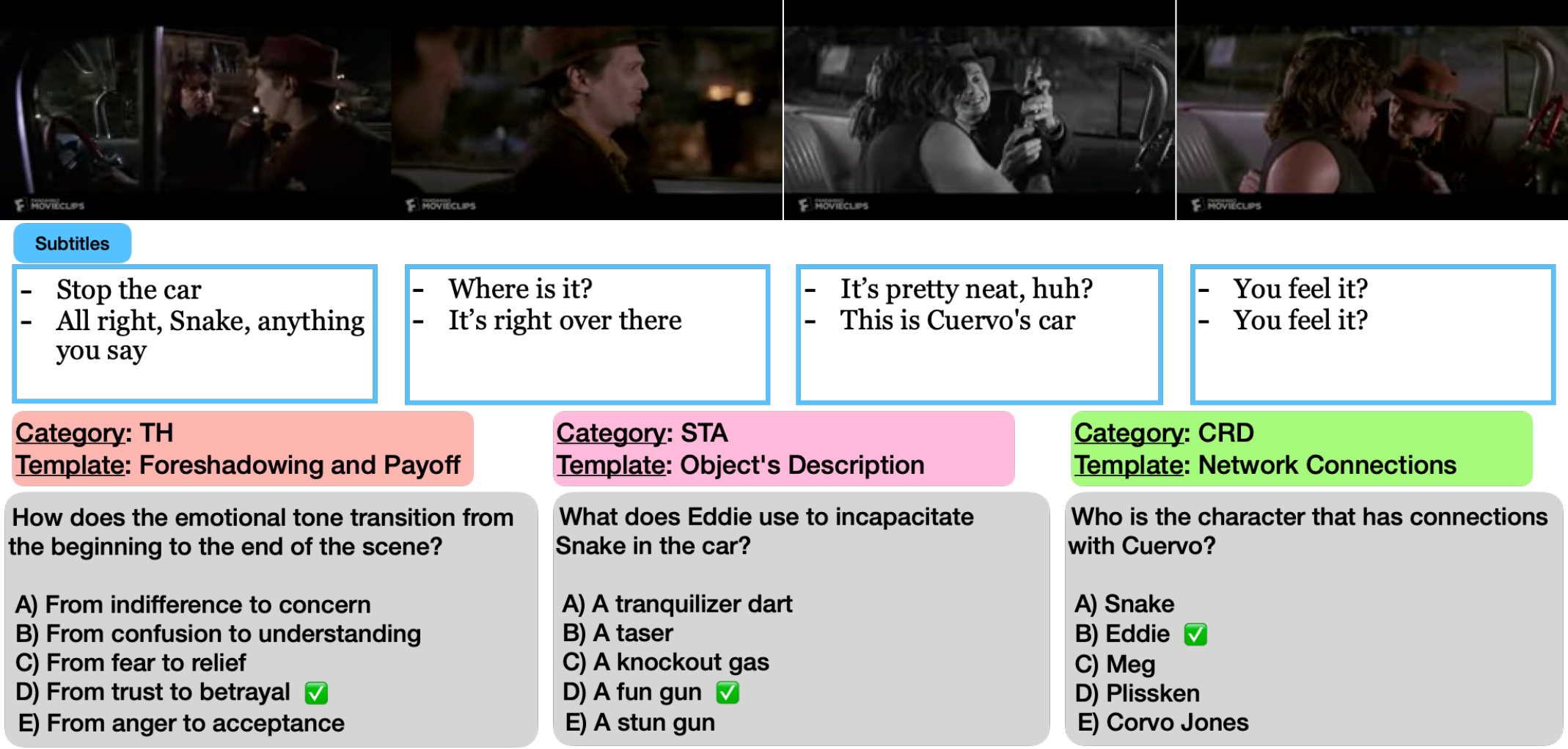}
        \caption{}
        \label{fig:scene_fig_3}
    \end{subfigure}
    
    \vspace{0.5cm}  
    
    \begin{subfigure}[b]{\textwidth}
        \includegraphics[width=\textwidth]{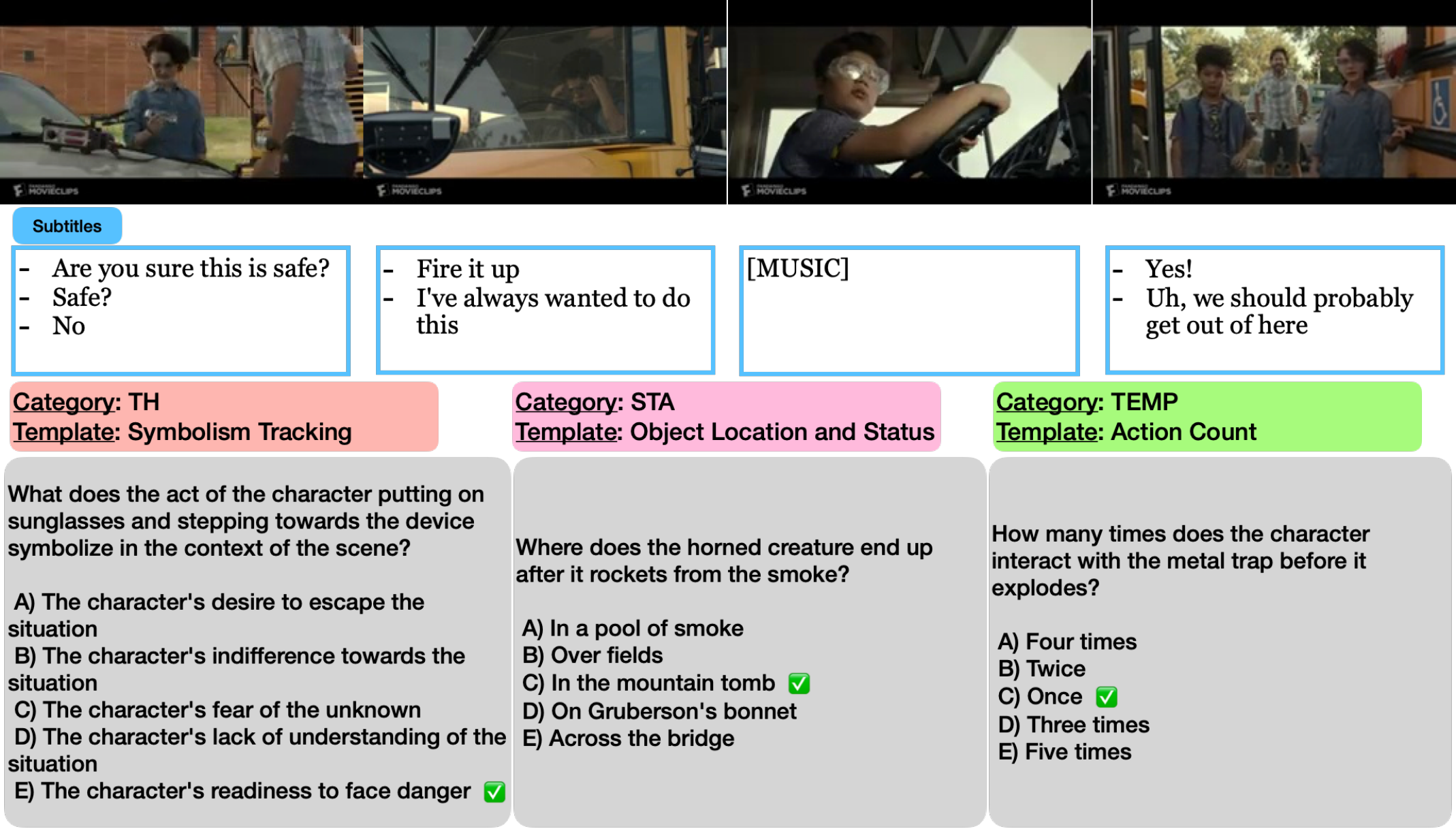}
        \caption{}
        \label{fig:scene_fig_4}
    \end{subfigure}
    
    \caption{\textbf{Example movie clip and multiple-choice questions from \logan}. The first and second rows depict a selection of image frames extracted from movie clips from  (a)\href{https://youtube.com/watch?v=aEIaR1nlEoo}{Escape From L.A.}, and  (b)\href{https://youtube.com/watch?v=0DDn8-m0QR0}{Ghostbusters: Afterlife}, accompanied by their corresponding subtitles. The next row showcases example questions along with the question template shown in colored headers. TEMP refers to Temporal. Please refer to \cref{tab:question_themes} for other acronyms.}
\end{figure}

\begin{figure}[htbp]
    \centering
    \begin{subfigure}[b]{\textwidth}
        \includegraphics[width=\textwidth]{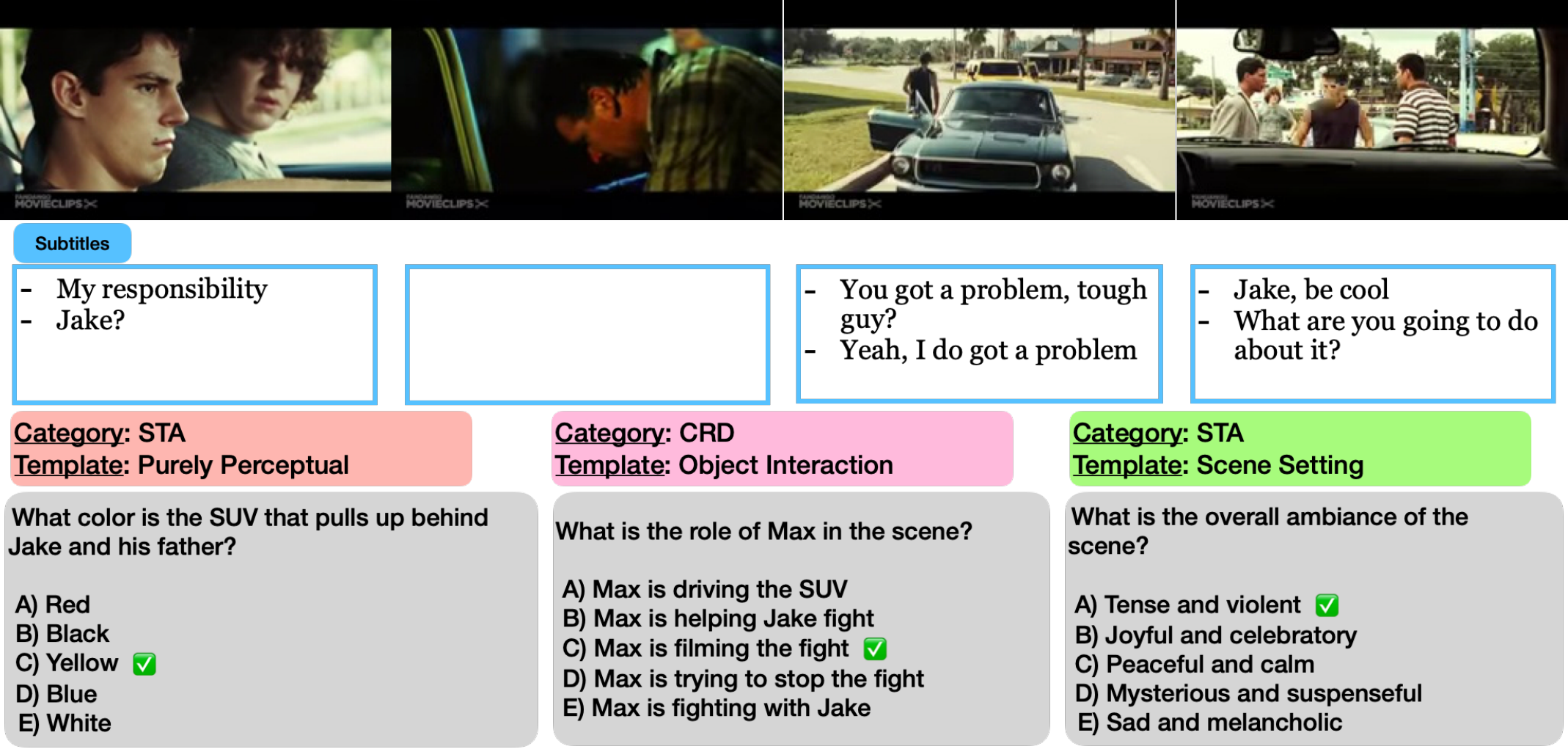}
        \caption{}
        \label{fig:scene_fig_5}
    \end{subfigure}
    
    \vspace{0.5cm}  
    
    \begin{subfigure}[b]{\textwidth}
        \includegraphics[width=\textwidth]{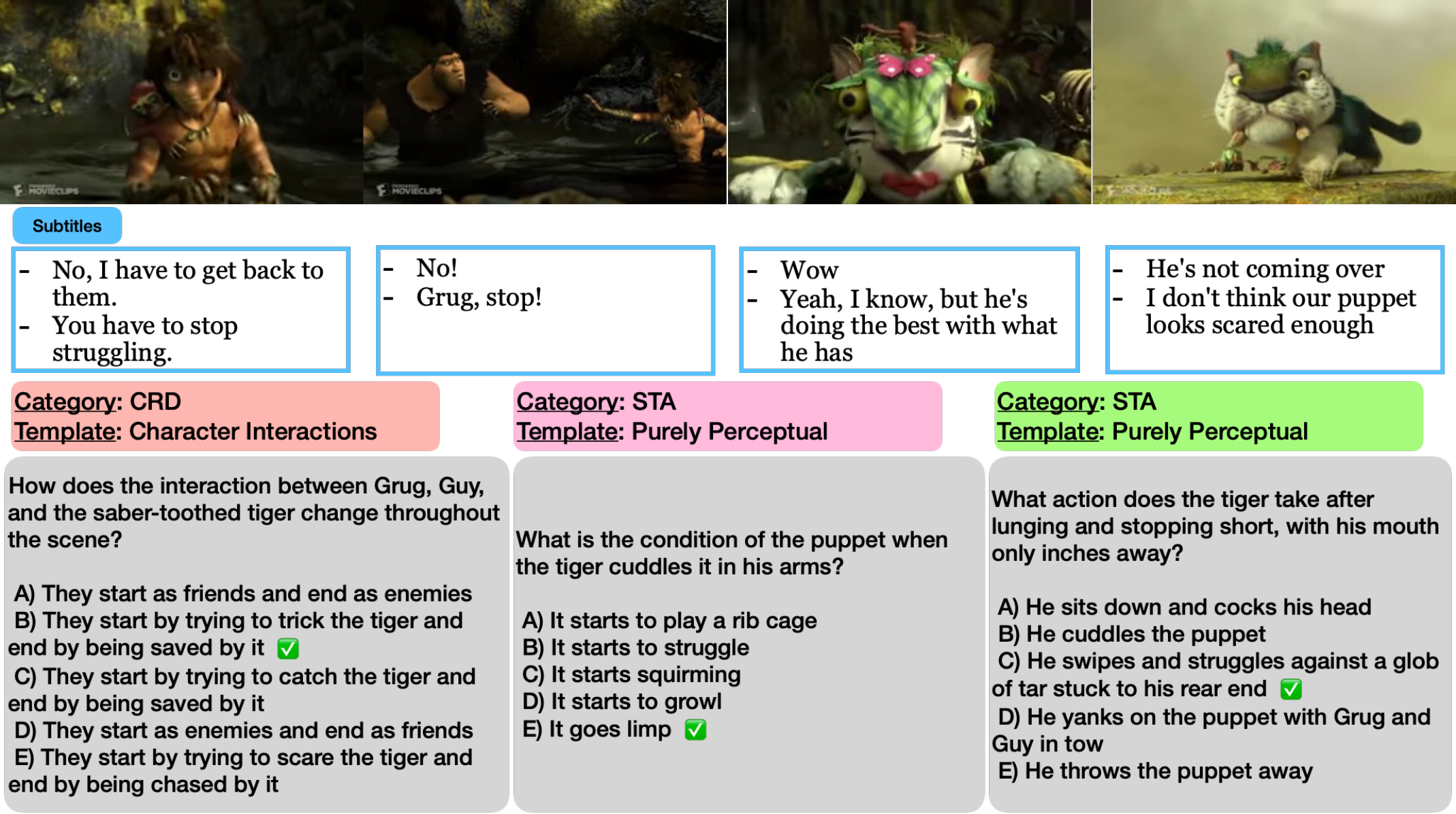}
        \caption{}
        \label{fig:scene_fig_6}
    \end{subfigure}
    
    \caption{\textbf{Example movie clip and multiple-choice questions from \logan}. The first and second rows depict a selection of image frames extracted from movie clips from (a) \href{https://youtube.com/watch?v=QQq6L1Sw4Ck}{Never Back Down}, and  (b) \href{https://youtube.com/watch?v=xxl1Hrw2eQM}{The Croods}, accompanied by their corresponding subtitles. The next row showcases example questions along with the question template shown in colored headers. TEMP refers to Temporal. Please refer to \cref{tab:question_themes} for other acronyms.}
\end{figure}

\begin{figure}[htbp]
    \centering
    \begin{subfigure}[b]{\textwidth}
        \includegraphics[width=\textwidth]{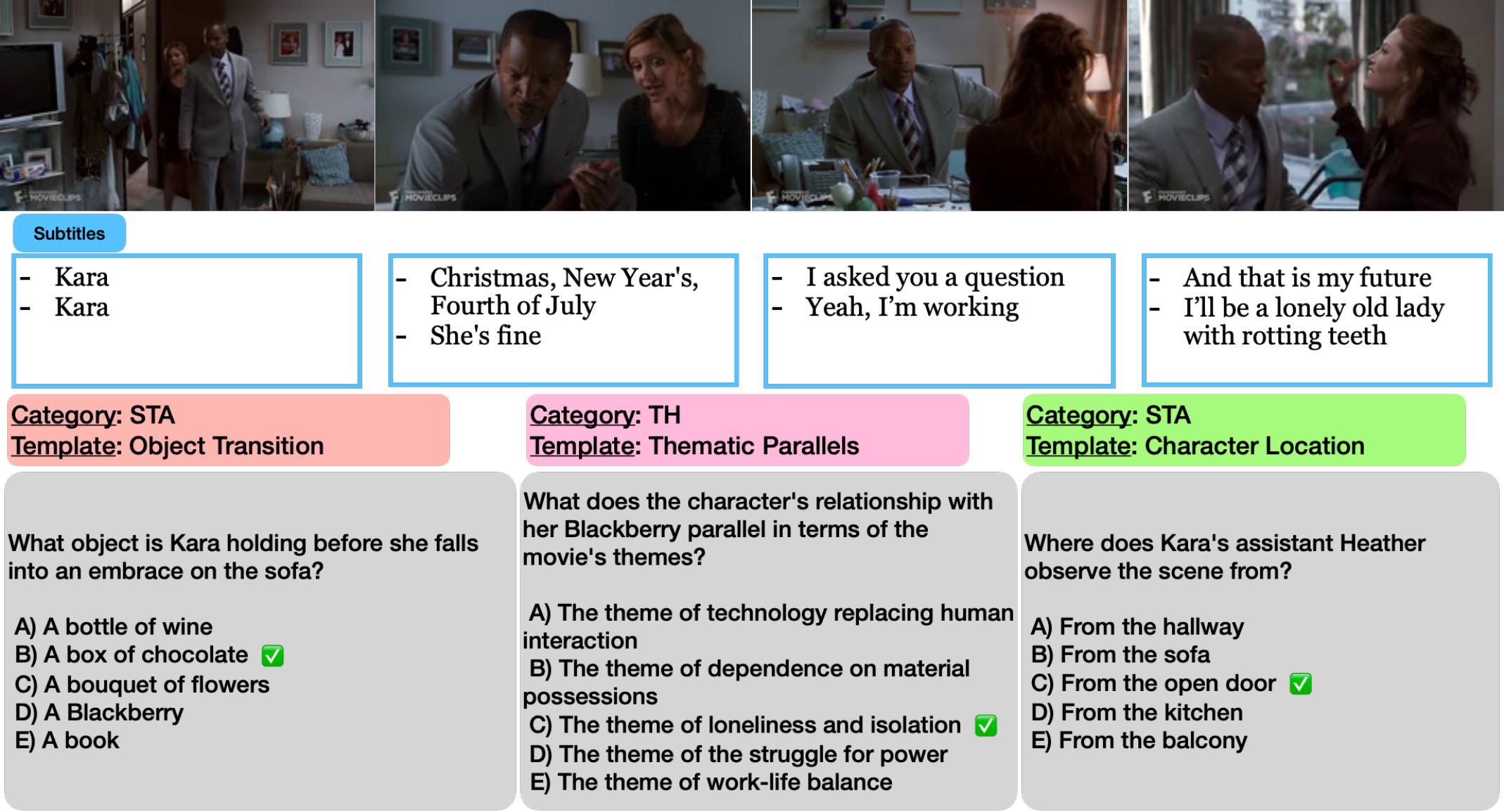}
        \caption{}
        \label{fig:scene_fig_7}
    \end{subfigure}
    
    \vspace{0.5cm}  
    
    \begin{subfigure}[b]{\textwidth}
        \includegraphics[width=\textwidth]{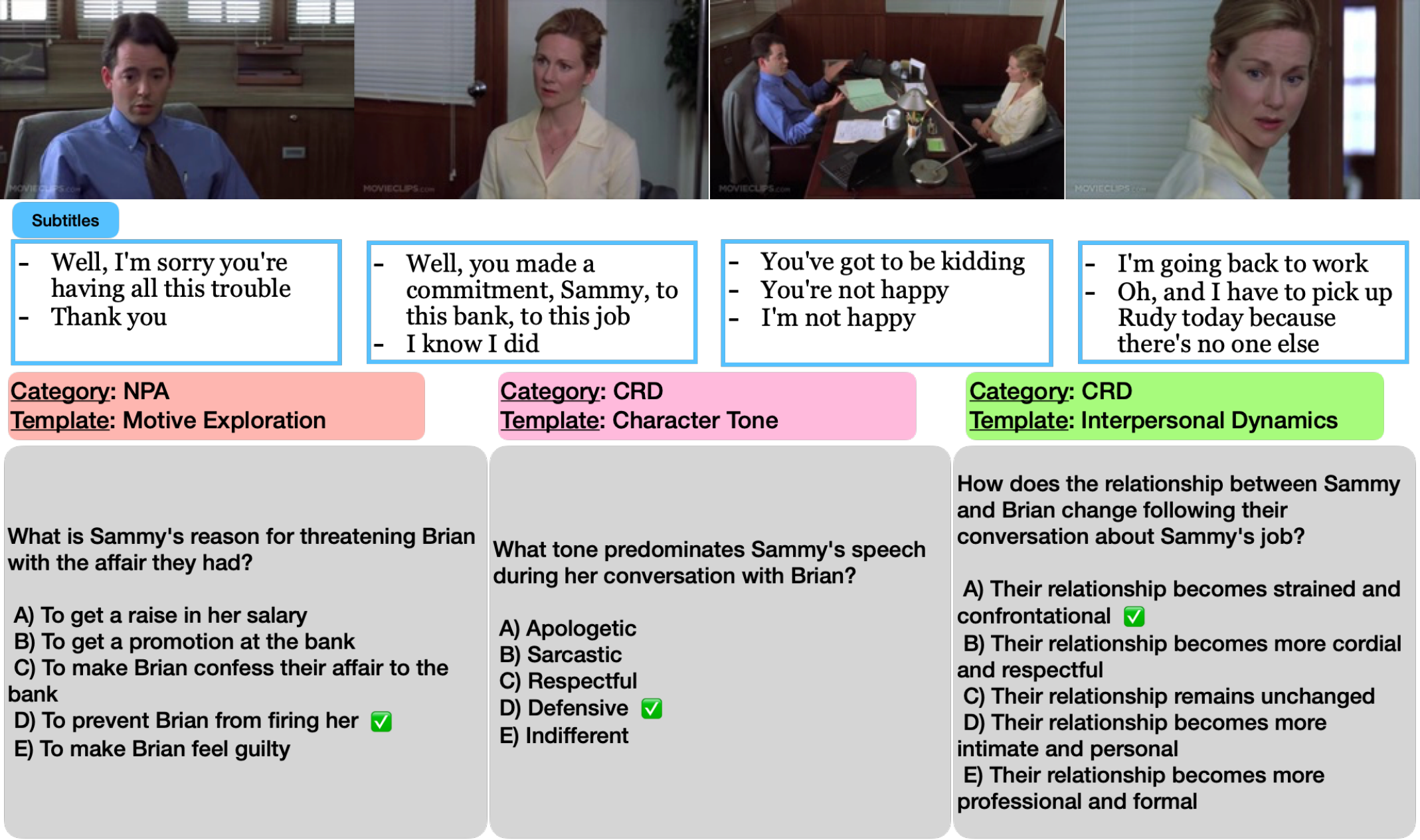}
        \caption{}
        \label{fig:scene_fig_8}
    \end{subfigure}
    
    \caption{\textbf{Example movie clip and multiple-choice questions from \logan}. The first and second rows depict a selection of image frames extracted from movie clips from (a) \href{https://youtube.com/watch?v=cmgeSY8YdO4}{Valentine's Day}, and (b) \href{https://youtube.com/watch?v=QX1qfAa0np8}{You Can Count on Me}, accompanied by their corresponding subtitles. The next row showcases example questions along with the question template shown in colored headers. TEMP refers to Temporal. Please refer to \cref{tab:question_themes} for other acronyms.}
\end{figure}

We present a few examples from our dataset in \cref{fig:scene_fig_1,fig:scene_fig_2,fig:scene_fig_3,fig:scene_fig_4,fig:scene_fig_5,fig:scene_fig_6,fig:scene_fig_7,fig:scene_fig_8}.

\section{Additional Related Work}
\label{addn_rel_work}
\textbf{Synthetic data with human in the loop.} Training models on synthetic data is a popular paradigm in recent times. We have seen many advances in generation as well as usage on synthetic data in recent times, both in vision~\cite{wood2021fake, bordes2024pug, tian2023learning, hemmat2023feedback} and language~\cite{alpaca, maini2024rephrasing, li2023textbooks, yuan2024self, wei2023simple}. For instance, Self-Instruct~\cite{wang2022self} proposes a pipeline to create an instruction dataset based on a few instruction examples and categories defined by humans. We mainly derived inspiration and the fact that modern LLMs are quite good at understanding long text and creating question-answer pairs. UltraChat~\cite{ding2023enhancing} is another synthetic language dataset which is created by using separate LLMs to iteratively generate opening dialogue lines, simulate user queries, and provide responses. This allows constructing large-scale multi-turn dialogue data without directly using existing internet data as prompts. Additionally, Evol-Instruct~\cite{xu2023wizardlm}, automatically generates a diverse corpus of open-domain instructions of varying complexities by prompting an LLM and applying iterative evolution operations like in-depth evolving (adding constraints, deepening, etc.) and in-breadth evolving (generating new instructions). To our knowledge, we are among the first to apply automated template generation and question synthesis techniques to vision and video modalities using LLMs.

\section{Additional QA Generation Details}
\label{appendix_sec:purely_perceptual}

In addition to the hand-crafted perceptual templates, we also create long-form question and answers based on a scene's visual summary. To achieve this, we first generate a visual summary of a video clip. Then, we prompt the model to create question-answers solely based on that summary.

We create a pure visual summary of the scene by using a vision LLM, similar to some of the recent works\cite{wang2023internvid,zhang2023simple}. First, we use a shot detection algorithm to pick the important frames\footnote{\url{https://www.scenedetect.com/}}, then we annotate each of these frames with Gemini vision API (\texttt{gemini-pro-vision}). We ablated many SOTA open-source vision LLMs such as Llava 1.5-13B~\cite{liu2023improved}, OtterHD~\cite{li2023otter}, mPlug-Owl~\cite{ye2023mplug} and MinGPT-4~\cite{zhu2023minigpt}, along with Gemini and GPT-4V (\texttt{GPT-4-1106-vision-preview}). While GPT-4V has high fidelity in terms of image captioning, it is quite expensive. Most of the open-source LLM captions are riddled with hallucinations. After qualitatively evaluating across many scenes, we found that Gemini's frame descriptions are reliable and they do not suffer too much from hallucination. Once we have frame-level descriptions, we then pass the concatenated text to Gemini text model \texttt{gemini-pro} and prompt it to produce a short descriptive summary of the whole scene.  Even though Gemini's scene visual summary is less likely to have hallucinated elements, we however spotted a few hallucinated sentences. Hence all the MCQs generated using this summary are added only to the training split but not to the eval split.

\section{Question Template Category Details}
\label{appendix_sec:qgen_details}

\textbf{Character and Relationship Dynamics:} This category would include templates that focus on the actions, motivations, and interactions of characters within the movie. It would also cover aspects such as character roles, reactions, decisions, and relationships.

\textbf{Narrative and Plot Analysis:} This category would encompass templates that delve into the storyline, plot twists, event sequences, and the overall narrative structure of the movie. It would also include templates that explore the cause-and-effect dynamics within the plot.

\textbf{Thematic Exploration:} This category would include templates that focus on the underlying themes, symbols, motifs, and subtext within the movie. It would also cover aspects such as moral dilemmas, emotional responses, and the impact of discoveries. 

\textbf{Setting and Technical Analysis:} This category would encompass templates that focus on the setting, environment, and technical aspects of the movie. It would include templates that analyze the location of characters and objects, the use of props, the impact of interactions on the environment, and the description and function of objects. 

\textbf{Temporal:} This category pertains to questions and answers that assess a model's comprehension of a movie clip's temporal aspects, such as the accurate counting of specific actions, the understanding of the sequence of events, etc.

\begin{table}[hbp]
    \centering
     \caption{Sample templates and prototypical questions from each of the categories}
     \vspace{0.15cm}
\resizebox{0.9\textwidth}{!}{
    \begin{tabular}{p{0.3\linewidth}| p{0.3\linewidth} | p{0.4\linewidth}} \toprule
    \textbf{Category} & \textbf{Question template} & \textbf{Prototypical question} \\ \midrule
    
        Character and Relationship Dynamics (CRD) &  Interpersonal Dynamics & What changes occur in the relationship between person A and person B following a shared experience or actions? \\ \midrule
         Character and Relationship Dynamics (CRD) & Decision Justification & What reasons did the character give for making their decision? \\\midrule
         Narrative and Plot Analysis (NPA) & Crisis Event & What major event leads to the character's drastic action? \\\midrule
         Narrative and Plot Analysis (NPA) & Mysteries Unveiled & What secret does character A reveal about event B? \\\midrule
         Setting and Technical Analysis (STA) & Physical Possessions & What is [Character Name] holding? \\
         \midrule
          Setting and Technical Analysis (STA) & Environmental Details & What does the [setting/location] look like [during/at] [specific time/place/event]? \\\midrule
         Temporal (TEMP) & Critical Time-Sensitive Actions & What must [Character] do quickly, and what are the consequences otherwise?\\
         \midrule
         Temporal (Temp) & Frequency & How many times does a character attempt [action A]?\\\midrule
         Thematic Exploration (TH) & Symbolism and Motif Tracking & Are there any symbols or motifs introduced in Scene A that reappear or evolve in Scene B, and what do they signify? \\\midrule
         Thematic Exploration (TH) & Thematic Parallels & What does the chaos in the scene parallel in terms of the movie's themes? \\
  \bottomrule
    \end{tabular}}
    \label{tab:question_themes}
\end{table}
\section{QA Generation by Different Models}
\label{appendix_sec:prop_model_ques_gen}

\begin{table}[!h]
\centering
\caption{Comparing question-answer pairs generated by GPT-4 with those generated by Gemini, for the movie clip: \href{https://www.youtube.com/watch?v=vsBwRV2b3LY}{The Heartbreak Kid (3/9) Movie CLIP - Taking the Plunge (2007) HD}. TEMP refers to Temporal. Please refer to \cref{tab:question_themes} for other acronyms.} 
\label{table:comparison_gpt_gemini_1} 
\scalebox{0.7}{
\begin{tabular}{p{1.2cm}|p{7cm}|p{7cm}}
\toprule
\textbf{Category} & \textbf{GPT-4 Generated QA} & \textbf{Gemini Generated QA} \\
\midrule
CRD & Question: What is the significant event that Eddie and Lila are celebrating? \newline 
- A) Their wedding \cmark \newline 
- B) Their first date anniversary \newline 
- C) Lila's birthday \newline 
- D) Their engagement \newline 
- E) Eddie's promotion at work  & Question: What is Eddie doing at the beginning of the scene? \newline 
- A) Dancing with Lila \newline 
- B) Giving a speech \newline 
- C) Cutting the wedding cake \newline 
- D) Kissing Lila \cmark \newline 
- E) Talking to his friends  \\
\midrule
NPA & Question: What incident leads to the main character's change in attitude towards marriage? \newline 
- A) His friend's advice \cmark \newline 
- B) His mother's arrival \newline 
- C) His bride's beauty \newline 
- D) His friend's gift \newline 
- E) His bride's dress  
 & Question: How does Eddie resolve his conflict with his friend? \newline 
- A) He apologizes for his past behavior. \newline 
- B) He confronts his friend about their differences. \newline 
- C) He ignores his friend and moves on. \newline 
- D) He seeks revenge on his friend. \newline 
- E) He reconciles with his friend. \cmark \\
\midrule
TEMP & Question: How long is the couple planning to take off for their road trip? \newline 
- A) One week \newline 
- B) Four weeks \newline 
- C) Five weeks \newline 
- D) Two weeks \newline 
- E) Three weeks \cmark
 & Question: What occurs immediately after the wedding ceremony? \newline 
- A) The couple kisses. \newline 
- B) The guests congratulate the couple. \newline 
- C) The bride's mother arrives. \cmark \newline 
- D) The couple leaves for their honeymoon. \newline 
- E) The groom gives a speech. \\
\midrule
STA & Question: Where is the gift Eddie's friend gives him supposed to end up? \newline 
- A) With Uncle Tito \cmark \newline 
- B) With Lila \newline 
- C) With Eddie \newline 
- D) With the wedding guests \newline 
- E) With Eddie's mom   & Question: What is the primary color of Lila's dress in the scene? \newline 
- A) Red \newline 
- B) Blue \newline 
- C) Yellow \newline 
- D) Green \newline 
- E) White \cmark \\
\midrule
TH & Question: How does the emotional tone shift from the beginning to the end of the scene? \newline 
- A) From excitement to disappointment \newline 
- B) From joy to sorrow \newline 
- C) From anticipation to regret \newline 
- D) From happiness to surprise \cmark \newline 
- E) From nervousness to relief 
& Question: What does the chaotic atmosphere at the reception symbolize in relation to the film's themes? \newline 
- A) The unpredictability of life \cmark \newline 
- B) The challenges of marriage \newline 
- C) The importance of family \newline 
- D) The power of love \newline 
- E) The fragility of relationships  \\
\bottomrule
\end{tabular}
}

\end{table}

In this section, we present example question-answer (QA) pairs generated by GPT-4 and Gemini across various question categories in Table~\ref{table:comparison_gpt_gemini_1} and Table~\ref{table:comparison_gpt_gemini_2}. As alluded to in the main paper, we note that GPT-4 consistently produces high-quality questions in all categories. In contrast, Gemini works well only for a few select categories, namely, Character Relationships and Interpersonal Dynamics (CDR), and Setting and Technical Analysis (STA). The gap in quality of the QA generated stems not only from the implicitly better and diverse concepts captured by GPT-4, but also from the hallucination tendencies of Gemini. For instance, in Table-~\ref{table:comparison_gpt_gemini_1}, Gemini mistakes the dialogue -- “Thank you for talking some sense into me, man”, between Eddie and his friend as a suggestion for conflict resolution, and forms a narrative question based on it -- “How does Eddie resolve his conflict with his friend?”. Similarly, in Table~\ref{table:comparison_gpt_gemini_2}, Gemini misremembers the temporal sequence and selects a wrong option as the answer choice for the temporal category. We quantify the quality of generated questions across the different choices of question-generation, and template selection models in Tab.~\ref{table:qagen_model_comparison_quant}. Here, we note that while the GPT-4 \& GPT-4 combination results in the fewest degenerate questions, the Gemini \& GPT-4 pairing also performs well and is cost-efficient on a large scale.

\begin{table}[!h]
\centering
\caption{Comparing question-answer pairs generated by GPT-4 with those generated by Gemini, for the movie clip: \href{https://www.youtube.com/watch?v=ZGFA2txwrg4}{Ghostbusters: Afterlife (2021) - Muncher Attack Scene (3/7) | Movieclips}. TEMP refers to Temporal. Please refer to \cref{tab:question_themes} for other acronyms.} 
\label{table:comparison_gpt_gemini_2} 
\scalebox{0.65}{
\begin{tabular}{p{1.2cm}|p{7cm}|p{7cm}}
\toprule
\textbf{Category} & \textbf{GPT-4 Generated QA} & \textbf{Gemini Generated QA} \\
\midrule
CRD &Question: How does Phoebe's interaction with the ghost change throughout the scene? \newline 
- A) She goes from hiding to attacking. \cmark \newline 
- B) She goes from attacking to hiding. \newline 
- C) She goes from running away to attacking. \newline 
- D) She goes from communicating to attacking. \newline 
- E) She goes from hiding to running away. & Question: How does the interaction between Phoebe and the ghost change throughout the scene? \newline 
- A) Phoebe becomes more aggressive \cmark \newline 
- B) The ghost becomes more passive \newline 
- C) Phoebe becomes more fearful \newline 
- D) The ghost becomes more aggressive \newline 
- E) They remain neutral towards each other \\
\midrule
NPA & Question: What major event prompts Phoebe to take drastic action against the ghost? \newline 
- A) The ghost scuttles off and pipes burst into flame. \newline 
- B) The ghost hides behind machinery. \newline 
- C) The ghost starts eating a piece of pipe. \newline 
- D) The ghost belches metal fragments that spark and ricochet around them. \cmark \newline 
- E) The ghost starts searching the ground. 
 & Question: What observation prompts Phoebe to take action? \newline 
- A) The ghost's fear of the Aztec death whistle \newline 
- B) The ghost's vulnerability to proton blasts \newline 
- C) The ghost's search for something on the ground. \cmark \newline 
- D) The ghost's reaction to Podcast's camera goggles \newline 
- E) The ghost's belching of metal fragments 
 \\
\midrule
TEMP & Question: What happens immediately after the ghost belches metal fragments? \newline 
- A) Phoebe ducks down. \newline 
- B) The ghost scuttles off and pipes burst into flame. \newline 
- C) Podcast blows the Aztec death whistle. \newline 
- D) Phoebe powers up and fires a steady stream of protons. \cmark \newline 
- E) Phoebe pokes her head up. 
 & Question: Between which two events does Phoebe duck down? \newline 
- A) The ghost searches the ground and Phoebe pokes her head up. \newline 
- B) The ghost chomps on a pipe and Phoebe pokes her head up. \newline 
- C) Podcast blows the whistle and the ghost belches metal fragments. \newline 
- D) The ghost scuttles off and pipes burst into flame. \cmark \newline 
- E) Phoebe fires protons and the ghost pokes its head out. 
 \\
\midrule
STA & Question: Where do Podcast and Phoebe hide during the ghost encounter? \newline 
- A) Inside a car \newline 
- B) In a building \newline 
- C) Behind a tree \newline 
- D) Under a table \newline 
- E) Behind machinery \cmark  & Question: What is the primary material of the object that the ghost is chewing on? \newline 
- A) Wood \newline 
- B) Metal \cmark \newline 
- C) Plastic \newline 
- D) Rubber \newline 
- E) Fabric  \\
\midrule
TH & How does the emotional tone shift throughout this scene? \newline 
- A) From calm to chaotic \newline 
- B) From fear to courage \cmark \newline 
- C) From confusion to understanding \newline 
- D) From excitement to disappointment \newline 
- E) From sadness to joy 
 & Question: How does the emotional tone shift from the characters' initial fear to their determination? \newline 
- A) The podcast's calmness inspires Phoebe to become more assertive. \newline 
- B) The ghost's search for something on the ground creates a sense of urgency. \newline 
- C) The characters' realization that they have a plan instills confidence. \cmark \newline 
- D) The ghost's belching of metal fragments intensifies the fear and chaos. \newline 
- E) The characters' decision to use the trap marks a shift from fear to determination.  \\
\bottomrule
\end{tabular}
}
\end{table}

\begin{table}[!h]
\centering
\caption{Comparison of Template Selection and Question Generation Models in generating better questions (lower degenerate questions) for a subset of movie clips. While the GPT-4 GPT-4 combination performs the best, Template Selection model has minimal effect.}\scalebox{0.9}{
\renewcommand{\arraystretch}{1} 
\label{table:qagen_model_comparison_quant}
\begin{tabular}{c|c|cc}
\toprule
\textbf{Template Selection Model} & \textbf{Question Generation Model} & \textbf{\% Degenerate Questions} \\
\midrule
Gemini & Gemini & 25.12 \\
Gemini & GPT-4 & 18.51 \\
GPT-4 & Gemini & 21.66 \\
GPT-4 & GPT-4 & 13.88 \\
\bottomrule
\end{tabular}
}
\vspace{2pt}

\end{table}

\section{Train data statistics}
We present the question category statistics of train split in \cref{appendix_fig:dataset_stats_train}.
\begin{figure}
     \centering
     \begin{subfigure}[b]{0.35\textwidth}
         \centering
         \includegraphics[width=\textwidth]{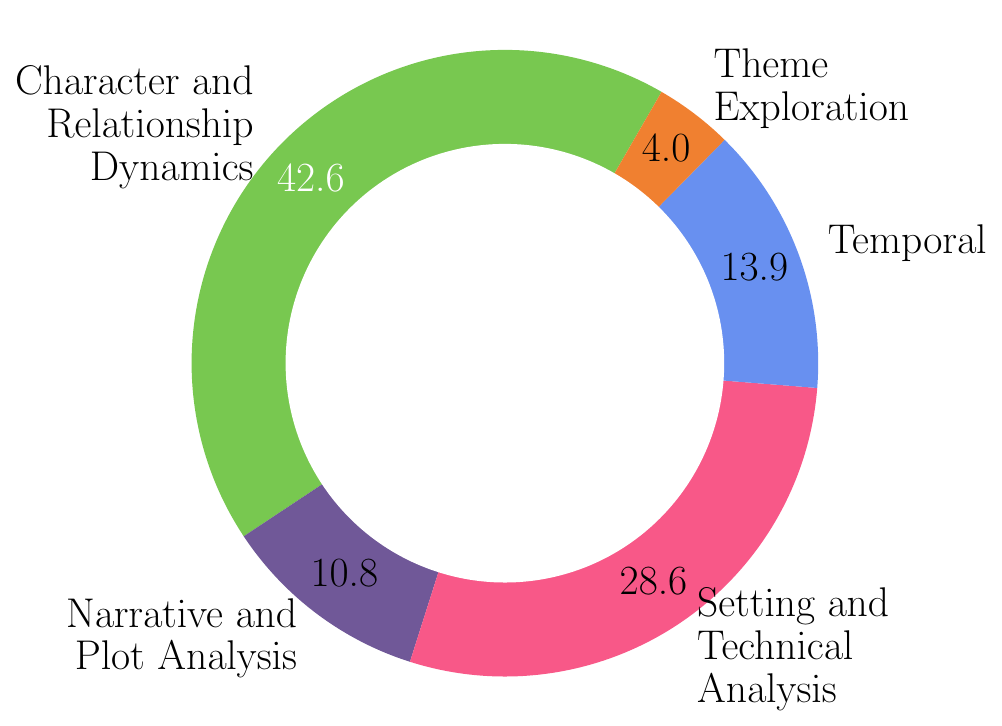}
         \label{appendix_fig:question_theme_distribution_train}
     \end{subfigure}

        \caption{Question category composition in the train split of the dataset. }
       \label{appendix_fig:dataset_stats_train}
\end{figure}

\section{Additional Evaluation Details}
\label{appendix_sec:addn_eval_details}
We use two NVIDIA A40 GPUs, each with 48GB of memory, and two NVIDIA A100, each with memory of 82GB, for experiments with open-source models. The model versions and dates are as follows: Gemini 1.5 Pro [gemini-1.5-pro-001] and Gemini 1.5 Flash [gemini-1.5-flash-001], from May 20th to June 1st, 28th. GPT-4o [gpt-4o-2024-05-13] was used on May 14th, 2024; GPT-4 Vision [gpt-4-turbo], Gemini Pro Vision [gemini-pro-vision], and Claude 3 (Opus) [claude-3-opus-20240229] were used from April 29th to May 10th, 2024. The Gemini 1.5 models throw safety-blocking exceptions for a few of the videos, hence we could only evaluate them on $\approx$ 4.2k samples out of 4941. The closed-source models in our evaluations (GPT-4, Gemini, Claude families) are released by their respective creators under proprietary licenses. In contrast, open-source models are released under various ope-source licenses such as CC BY-NC-SA 4.0, BSD 3-Clause ``New" or ``Revised" License, etc.
\section{Additional Evaluation Strategies}
\label{appendix_sec:additional_metrics}
\begin{table}[!h]
\centering
\caption{Analyzing raw responses generated by OSS models, scores assigned by our evaluation pipeline, and corresponding failure modes for the movie clip: \href{https://www.youtube.com/watch?v=duU5cdQtpSE}{Area 51 (2015) - Sneaking Onto the Base Scene (4/10)}} 
\label{table:oss_raw_response}
\vspace{0.15cm}
\scalebox{0.7}{
\begin{tabular}{p{5cm}|p{1.5cm}|p{3cm}|p{1.5cm}|p{3cm}}
\toprule
\textbf{Question} & \textbf{Answer\newline Key} & \textbf{Model\newline Response} & \textbf{Assigned\newline Score} & \textbf{Failure\newline Mode} \\
\midrule
How does the emotional tone transition during the scene?\newline
- A) From despair to hope\newline
- B) From fear to acceptance\newline
- C) From confusion to understanding\newline
- D) From tension to panic\newline
- E) From anxiety to excitement & From anxiety to excitement & \texttt{“5, 4, 3, 2, 1.\textbackslash n            All right, I think I got it, guys.\textbackslash n   You did it?\textbackslash n
…..\newline
…..\newline
…..\newline This is the base.”}  & 0 & The model doesn’t choose any of the options, rather just repeats all the subtitles. \\
\midrule
What action does Darren take after observing the broken sensor?\newline
- A) Calls for help\newline
- B) Repairs it\newline
- C) Panics\newline
- D) Ignores it\newline
- E) Suggests next steps & Suggests next steps & \texttt{"\textbackslash n  ) Darren calls for help.”} & 0 & While we are able to reliably extract the choice present, it doesn’t match the answer key.\\
\midrule
Where is Jelena in relation to the helicopter at the end of the scene?\newline
- A) Inside\newline
- B) In front\newline
- C) To the side\newline
- D) Directly underneath\newline
- E) On it&Directly underneath& \texttt{Jelena is standing in front of the helicopter at the end of the scene.}&0&While we are able to reliably extract the choice present, it doesn’t match the answer key.\\
\midrule
How do Darren and Reed first interact with each other in the scene?\newline
- A) They argue\newline
- B) They fight\newline
- C) They exchange information\newline
- D) They greet each other\newline
- E) They ignore each other&They exchange information&\texttt{In the scene, Darren and Reed are seen walking together, and they exchange information with each other.} & 1 & We are able to extract the response, and it matches the answer key.\\
\bottomrule
\end{tabular}
}
\end{table}

\begin{table}[htp]
\centering
\caption{Performance of various models on \logan’s test split, as evaluated using various video captioning metrics -- BERTSCoRE~\citep{devlin2018bert}, CIDEr~\citep{vedantam2015cider}, ROUGE-L~\citep{lin2004rouge}.}
\label{tab:caption_eval}
\scalebox{0.8}{
\renewcommand{\arraystretch}{1.2}
\begin{tabular}{c|cccc}
\toprule
Model & BERTScore$\uparrow$ & CIDEr$\uparrow$ & ROUGE-L$\uparrow$ \\
\midrule
mPLUG-Owl~\cite{ye2023mplugowl} & 0.38 & 0.74 & 0.22 \\
Video-ChatGPT~\cite{Maaz2023VideoChatGPT} & 0.39 &  0.63 & 0.23  \\
Intern-VL-2 (1B) ~\cite{song2023moviechat} & 0.40 & 1.33 & 0.28  \\
CogVLM-2 ~\cite{song2023moviechat} & 0.45 & 1.20 & 0.31  \\
\bottomrule
\end{tabular}
}
\end{table}
As discussed in Sec.~\ref{sec:eval_sec} of the main paper, we evaluate a model’s performance on \logan’s test-split by computing its accuracy in choosing the correct answer from a set of multiple-choice options. This involves extracting the chosen answer from the model's raw response and mapping it to one of the predefined answer options. While we perform extensive prompt tuning to ensure the model outputs only the option-letter in its response and rigorously post-process responses to separately extract the chosen option-letter and the corresponding option-text generated (if generated), there remains a possibility of errors. The model may not always follow these instructions perfectly and could produce verbose responses with unnecessary text snippets, such as "In my opinion," "The correct answer is," or "... is the correct answer."

Therefore, in this section, we compute traditional video-caption evaluation metrics that emphasize the semantic similarity between the answer key text and the raw model response, instead of exact string matching. We focus our evaluation and discussion on open-source models here, as we qualitatively noted that proprietary models, such as GPT-4V, Gemini-Pro, and Claude, strictly adhere to the prompt instructions, producing only the option letter in their response.
Specifically, we calculate the following video-captioning metrics -- BERTScore~\citep{zhang2019bertscore}, CIDEr~\citep{vedantam2015cider}, and ROUGE-L~\citep{lin2004rouge}. BERTScore calculates the contextual similarity between the answer key and model response in the embedding space of a pretrained transformer model like BERT-Base. Calculating the similarity between the latent representations, instead of direct string matching, provides robustness to paraphrasing differences in the answer key and model response. In contrast, CIDEr evaluates the degree to which the model response aligns with the consensus of a set of reference answer keys. In our setup, each question is associated with only one reference answer. The alignment here is computed by measuring the similarity between the non-trivial n-grams present in the model response and the answer key. Finally, ROUGE-L computes the similarity between the answer key and model response based on their longest common subsequence.  

We evaluate four open source models, i.e. mPLUG-Owl, Video-ChatGPT, Intern-VL-2 (1B), and CogVLM2, using the aforementioned metrics and report the results in Table~\ref{tab:caption_eval}. In line with the accuracy trend in the main paper. These findings further support the reliability of our normalization and post-processing steps during accuracy computation.
\section{Human Study Details}
\label{appendix:human_study_details}
\begin{figure}[h]
    \centering
    \includegraphics[width=\textwidth]{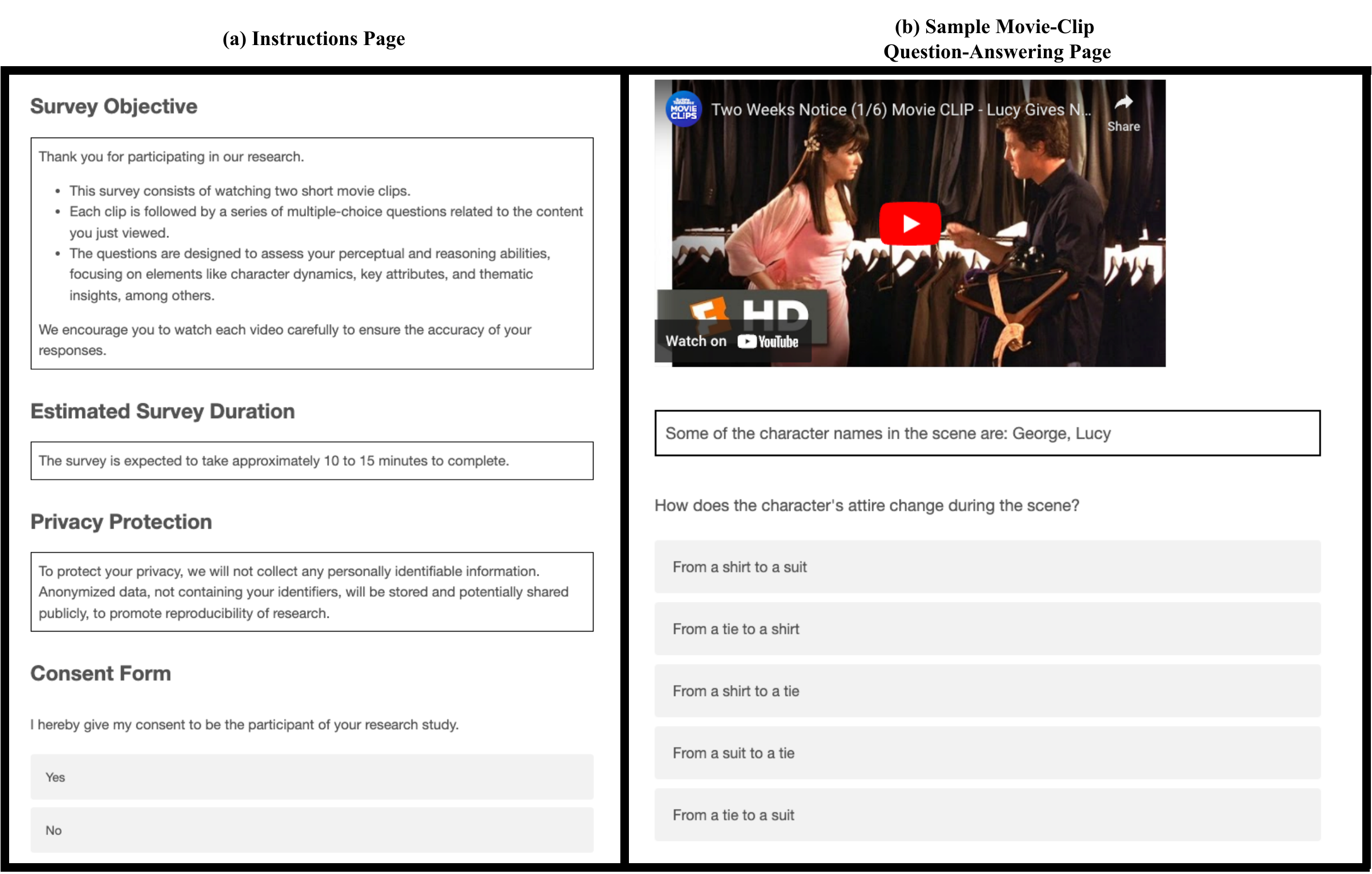}
    \caption{(\textit{left}) (a) \textbf{Instructions Page:} The instructions page at the beginning of the survey, as presented to participants. The participants provide informed consent before viewing any video clip and answering questions. (\textit{right}) (b) \textbf{Sample Movie-Clip Question-Answering Page:} An example of one of the movie clips and corresponding question, as presented to the participants. The participants are required to watch the clip and answer the questions by selecting the correct answer choice out of five options.}
    \label{fig:human_survey}
\end{figure}

\begin{figure}[tb]
  \centering
   \includegraphics[width=\textwidth]{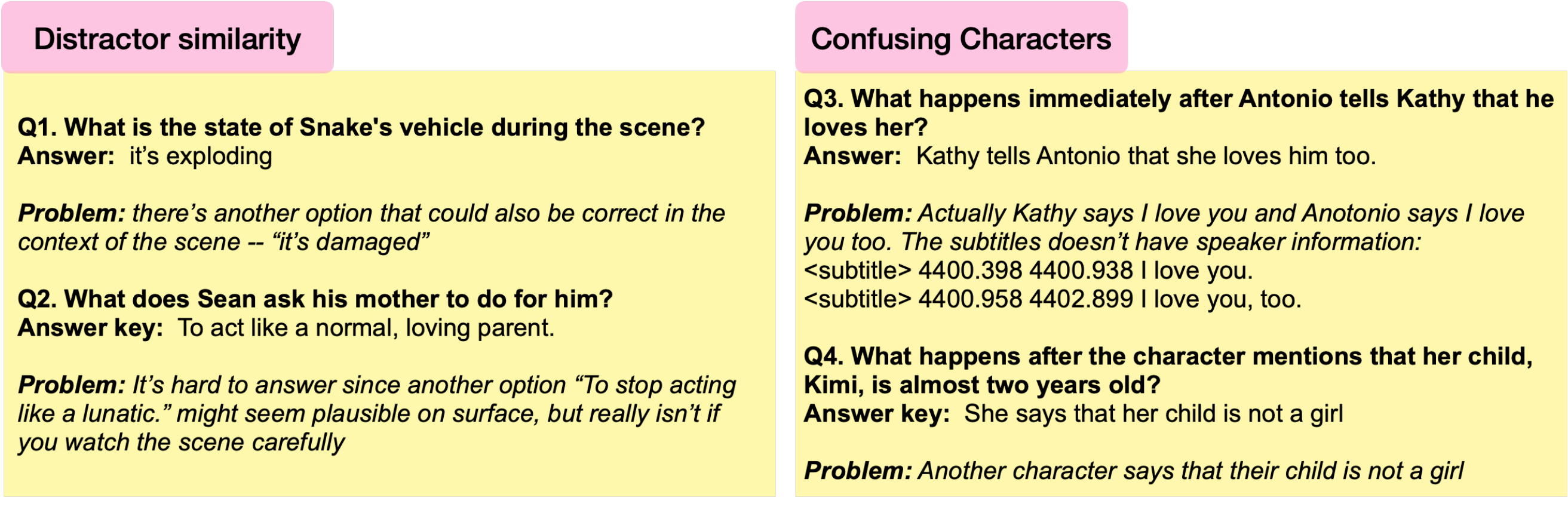}
  \caption{\textbf{Sample failure cases from human study}: We conducted a human study to check the quality of questions and we found a few systemic issues. We fixed all systemic issues in the final version of the dataset. The movie clip for Q1 can be found \href{https://www.youtube.com/watch?v=40p6dkKil_8}{here}; for Q2, \href{https://www.youtube.com/watch?v=DnwnDFr9kOs}{here}; for Q3, \href{https://www.youtube.com/watch?v=gknfkz5a-YQ}{here}; and for Q4, \href{https://www.youtube.com/watch?v=QizNYqfYekk}{here}. 
  }
  \label{fig:qa_failure_cases}
\end{figure}

\begin{figure}[tb]
  \centering
   \includegraphics[width=\textwidth]{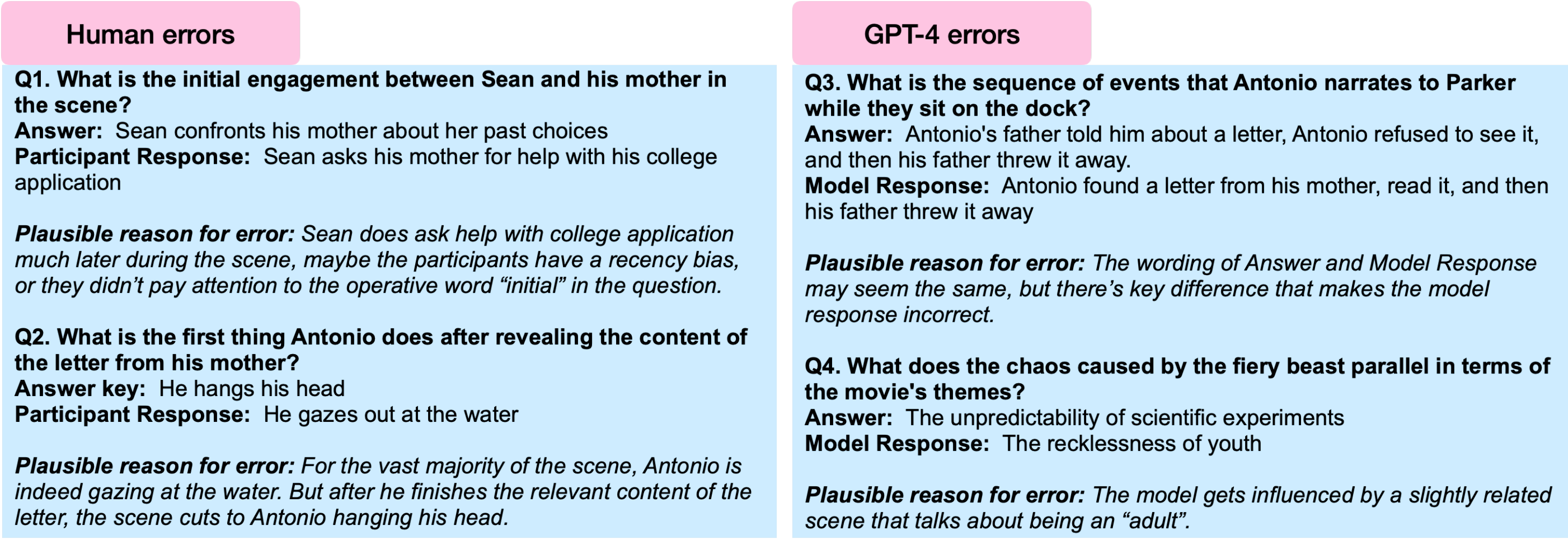}
  \caption{\textbf{Hard questions according to humans and \gpt V}: After conducting the human study, we looked at the questions which human got wrong and the questions which GPT-4 got wrong. Some of these questions are difficult and can only be answered by paying careful attention to the video. The movie clip for Q1 can be found \href{https://youtube.com/watch?v=DnwnDFr9kOs}{here}; for Q2 and Q3, \href{https://youtube.com/watch?v=ZqePSEpN56o}{here}; and for Q4, \href{https://youtube.com/watch?v=0DDn8-m0QR0}{here}. 
  }
  \label{fig:hard_qa_humans_gpt4}
\end{figure}

The authors conducted a small human study with 25 graduate student volunteers to evaluate the quality of the \logan dataset questions. Each participant answered ten randomly sampled multiple-choice questions about two video clips. Our human study survey was granted an exemption by our institute’s Institutional Review Board (IRB), and all participants gave their informed consent before viewing the videos and responding to the questions. For full instructions and consent questions given to participants, please refer to Fig.~\ref{fig:human_survey}-(a). Additionally, we did not collect any personally identifiable information from the participants.
It's important to note that our dataset consists of English movies produced in the United States. These films are likely certified by the Motion Picture Association of America (MPAA), which means they adhere to strict content standards and classification guidelines. As a result, they're expected to contain minimal offensive content.
An example of the question-answering page can be found in Fig.~\ref{fig:human_survey}-(b).

Post the study, we interviewed each participant after the survey to ask if they found any systematic issues in any of the questions they were asked to answer about the video. Later, a panel of authors audited all questions where humans got the answer wrong. We noticed that most of the time when a human got a question wrong it was likely due to one of the following reasons (i) due to their inability to attend over the entire clip at once, (ii) due to their inability to understand the dialogue or understand cultural references (iii) carelessness in answering,  as the correct answer was indeed present in the video. We did notice some problematic patterns with a small subset of questions. The main issue is distractor similarity, where humans found two plausible answers and they chose one randomly. We present a few such examples in \cref{fig:qa_failure_cases}.  We removed the questions from the test set for which we found ambiguous answers.

We again conducted a second human study on the test set's final version, and the human accuracy is 73\%. The authors have independently taken the survey, and the corresponding accuracy is 86\%. Once again, a careful investigation by a team of authors indicates that even most of these wrong answers are due to human error and confusion over the many events in a scene. We conclude from this study that many of the questions are answerable but difficult. We present the question category-level performance in Sec.~\ref{sec:eval_sec} in the main paper.

\section{Example Degenerate Questions}

\label{appendix_sec:degenerate_questions}
\begin{table}[!h]
\centering
\caption{\textbf{Example degenerate questions.} Examples of degenerate questions filtered from \logan. These questions can be categorized as degenerate for various reasons, including: being answerable through common sense (rows one to three) and the models possibly memorizing the movie scripts (rows four and five)}
\label{tab:degen_ques}
\scalebox{0.9}{
\begin{tabular}{m{0.3\textwidth}|m{0.7\textwidth}}
\toprule
Movie Clip &  Degenerate Questions \\
\midrule
\href{https://www.youtube.com/watch?v=v1M3w_o7cOc}{Scream (1996) - Wrong Answer Scene (2/12) | Movieclips} & Question: Where does the conversation between the characters take place? \newline 
- A) In a restaurant \newline 
- B) In a car \newline 
- C) In a classroom \newline 
- D) At a party \newline 
- E) Over the phone \cmark \\
\midrule
\href{https://www.youtube.com/watch?v=N5EYRBPqrgs}{The Godfather: Part 3 (8/10) Movie CLIP - Michael Apologizes to Kay (1990) HD} & Question: What thematic element is paralleled in the character's dialogue about his past and his destiny? \newline
- A) The theme of revenge \newline
- B) The theme of fate and free will \cmark \newline
- C) The theme of betrayal \newline
- D) The theme of lost innocence \newline
- E) The theme of love and sacrifice \\
\midrule
\href{https://www.youtube.com/watch?v=BM29Ze3d_cs}{The Croods (2013) - Try This On For Size Scene (6/10) | Movieclips} & Question: What happens right before Grug slips on a banana? \newline 
- A) Sandy helps Guy hand bananas out to all the monkeys. \newline 
- B) The saber-toothed cat roars at them from the bottom of a gorge. \newline 
- C) Grug throws a banana down angrily. \cmark \newline 
- D) Grug puts up his dukes and so does the monkey. \newline 
- E) Guy gives Grug a banana. \\
\midrule
\href{https://www.youtube.com/watch?v=flgiS8o13Eo}{Rugrats in Paris (2000) - We're Going to France! Scene (1/10) | Movieclips} & Question: What event prompts Kira Watanabe to call Mr. Pickles? \newline 
- A) The robot's destruction of the village. \newline 
- B) The robot's popularity among the villagers. \newline 
- C) The malfunction of the giant robot. \cmark \newline 
- D) The villagers' protest against the robot. \newline 
- E) The robot's successful performance. \\
\midrule
\href{https://www.youtube.com/watch?v=vLXjWGI8sfw}{Bottle Rocket (3/8) Movie CLIP - Future Man and Stacy (1996) HD} & Question: What happens immediately after Anthony and Dignan finish eating their sandwiches on the patio? \newline 
- A) Anthony chews a nut. \newline 
- B) A guy in a brown shirt approaches them. \cmark \newline 
- C) Stacey Sinclair introduces herself. \newline 
- D) Anthony tells his story about the beach house. \newline 
- E) Anthony goes to clean the pool. \\ 

\bottomrule
\end{tabular}
}
\end{table}

As discussed in Section~\ref{subsec:dataset_filtering} of the main paper, most question-answers generated are well-formed and include challenging distractors. However, a small minority are degenerate in that they can be answered directly, i.e., without viewing the movie video clip. To automatically filter out such questions, we formulate a degeneracy criterion. If a question can be answered by a wide variety of models without any context—that is, all models select the correct answer merely by processing the question and the five options—we label it as a degenerate question.
In this section, we present and discuss some of these degenerate questions in Table~\ref{tab:degen_ques}. We note that a question can be categorized as degenerate due to multiple possible reasons. For instance, consider the questions, “Where does the conversation between the characters take place?”, and “What happens right before Grug slips on a banana?”. The answer key for these corresponds to the most common-sense response, and the models are able to reliably identify the correct choices ("Over the phone", "Grug angrily throws a banana down") from among the distractions. There's another type of question that models might answer correctly if they've memorized the movie script. For example, the question, "What event prompts Kira Watanabe to call Mr. Pickles?" from the movie Rugrats in Paris, is accurately answered. This likely happens because of the memorization of the script and the distinct character names mentioned in the question.
\section{Additional Evaluation Results}

\subsection{Frame Rate Ablation}
\label{appendix_sec:frame_rate_ablation}
In this section we perform an ablation to investigate the utility of visual frames (from a model's perspective) by completely remove the visual frames and experiment solely with the provided dialogue when evaluating Video-LLMs. We do exactly this in Table~\ref{tab:sub_eval}, and observe that for all models, except Video-ChatGPT, performance significantly declines when evaluated with "only subtitles." This effect is more pronounced in commercial models compared to open-source ones. It appears that better overall models also tend to utilize visual information more effectively.

\begin{table}[htp]
\centering
\caption{Performance of models with video and subtitles (base case), and when only with subtitles on a subset of CinePile. TEMP - Temporal, CRD - Character and Relationship Dynamics, NPA - Narrative and Plot Analysis, STA - Setting and Technical Analysis, TH - Thematic Exploration.}
\label{tab:sub_eval}
\vspace{0.15cm}
\scalebox{0.9}{
\renewcommand{\arraystretch}{1} 
\begin{tabular}{c|c|cccccc}
\toprule
\textbf{Model} & \textbf{Average} & \textbf{CRD} & \textbf{NPA} & \textbf{STA} & \textbf{TEMP} & \textbf{TH} \\
\midrule
Gemini 1.5 Pro & 51.72 &  51.61 & 56.25 & 55.45 & 40.62 & 50.00  \\ 
(Only Subtitles) Gemini 1.5 Pro & 34.53 & 35.87 & 44.44 & 31.35 & 32.60 & 36.36  \\
\midrule

\midrule
GPT-4o & 50.45 & 51.14 & 66.66 & 52.54 & 34.78 & 45.45  \\ 
(Only Subtitles) GPT-4o & 37.23 & 45.03 & 44.44 & 29.66 & 28.26  & 45.45 \\
\midrule

\midrule
Video-LLaMA2 & 38.44 & 45.80 &  40.74 & 36.44 & 19.56 & 54.54  \\ 
(Only Subtitles) Video-LLaMA2 & 33.33 & 41.22 & 40.74 & 27.11 & 17.39 & 45.45 \\
\midrule

\midrule
Video-ChatGPT &  12.92 & 16.80 & 3.70 & 12.82 & 6.52 & 20.00 \\ 
(Only Subtitles) Video-ChatGPT & 16.16 & 22.04 & 11.53 & 12.71 & 13.04  & 9.09 \\
\midrule
 
\bottomrule
\end{tabular}
}
\end{table}

\subsection{Performance on Hard-Split}
\label{appendix_sec:hard_split_perf_cat}

\begin{figure}[!h]
    \centering
    \includegraphics[width=1.1\linewidth]{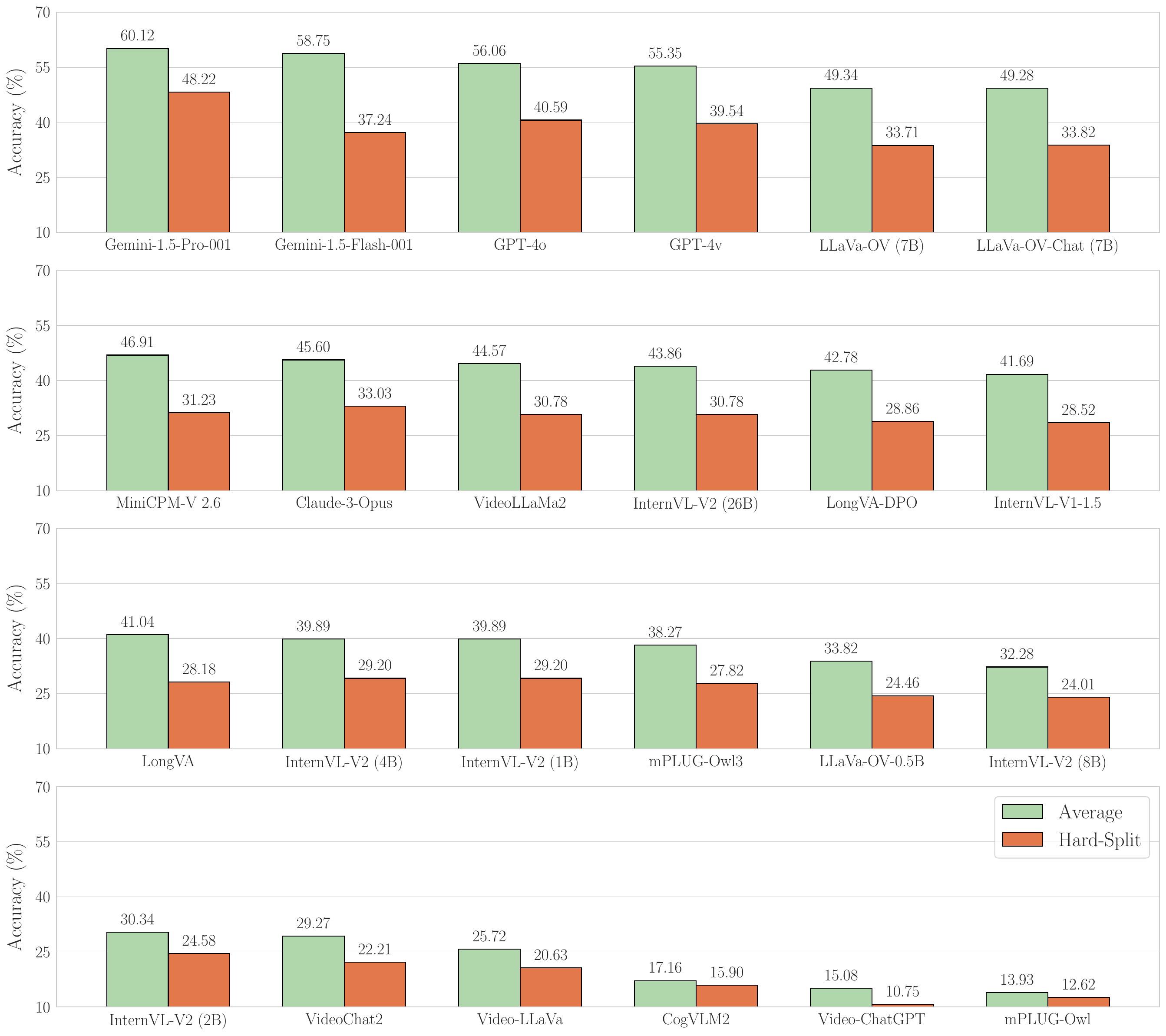}
    \caption{Models' performance on \logan test split, all questions vs hard questions.}
    \label{fig:hard_eval_all}
\end{figure}
\section{QA Generation Prompt}
\label{appendix_sec:qa_prompt}

\begin{quote}
As the curator of an advanced cinema analysis quiz, your expertise lies in designing intricate and diverse multiple-choice questions with corresponding answers that span the entire spectrum of film analysis.

- \textbf{Objective:} Create diverse and challenging questions based on the film analysis spectrum templates provided below. This spectrum is divided into five subcategories, each comprising several templates. Each template includes a title and a corresponding prototypical question or guideline. Avoid directly replicating the template title and these prototypical questions. Instead, your questions should reflect these elements' essence, even if not explicitly using the category titles in the question's wording.

\textbf{Mandatory Guidelines:}\\
  - \textbf{Template Use:} Use the provided question templates as a strict guide, ensuring that your questions are both relevant to the scene and varied in their analytical perspective. The prototype question in each template is for inspiration and should not be copied. Your questions should subtly reflect the prototype's essence, tailored to the specifics of the scene.
  
  - \textbf{Sub-Category Balance:} Ensure to generate an equal number of questions from each subcategory. This balance is crucial to cover a wide range of analytical perspectives.

- \textbf{Question and Answer Format:}\\
  - \textbf{Selected Template:} Indicate the film analysis Sub-Category and corresponding template your question is inspired by, without restricting the question's phrasing to the template's title.
  
  - \textbf{Questions:} Limited to one or two lines, formulated to be insightful and not overtly indicative of the answer. Avoid using direct template titles or overly descriptive language that could hint at the correct answer.
  
  - \textbf{Answers:} Five options per question, formatted as "\textbf{- A)}, \textbf{- B)}, \textbf{- C)}, \textbf{- D)}, and \textbf{- E)}", concise and reflective of the question's depth.
  
  - \textbf{Answer Key:} Specify the correct answer clearly with the formatting, "\textbf{Correct Answer:}", in the line following all the answer options.
  
  - \textbf{Rationale:} Write a rationale explaining the correctness of the "Answer Key" based on the scene's context in the next line.

\textbf{Input Information Format:}\\
- Movie scene details will be provided in a structured format comprising two distinct categories, and the relevant scene description. The two categories are as follows:

  - \textbf{\textless subtitle\textgreater} for character dialogues (to be used only for identifying character presence, not actions or dialogue content).
  
  - \textbf{\textless visual descriptions\textgreater} for noting characters' presence, attributes, thematic elements, etc., within the scene.

\textbf{Movie Scene:}
\{MOVIE\_SCENE\_TS\}

- \textbf{Spectrum of Film Analysis with Templates:} \\
Sub-Category: Character Analysis \\
\{TEMPLATES\_CHAR\}

Sub-Category: Narrative Understanding \\
\{TEMPLATES\_NARV\}

Sub-Category: Scene Setting \\
\{TEMPLATES\_SETTING\}

Sub-Category: Temporal \\
\{TEMPLATES\_TEMPORAL\}

Sub-Category: Theme \\
\{TEMPLATES\_THEME\}

\textbf{Instructions:}
Your task is to generate clear, unique, and insightful question-answer pairs strictly following the provided templates. Ensure the distribution of questions covers all subcategories evenly. Strictly avoid using words in the questions that give a strong hint about the answer. You can achieve this by keeping the questions concise and not using too many adjectives or adverbs in the question. Incorrect answers must be plausible and closely mirror the correct answer in length and form. The correct answer should not be deducible solely from the question and/or the wrong answers. After presenting all the options, the correct answer must be distinctly specified, but separate from the list of choices. Additionally, provide a concise rationale about why the question-answer falls into one of the selected templates from the Spectrum of Film Analysis by giving verbatim evidence from the subtitles and/or visual descriptions in the movie scene information.
\end{quote}

\newpage

\end{document}